\documentclass[twoside]{article}
\usepackage[accepted]{aistats2021}

\usepackage[utf8]{inputenc} % allow utf-8 input
\usepackage[T1]{fontenc}    % use 8-bit T1 fonts
\usepackage{lmodern}

\usepackage[round]{natbib}
\renewcommand{\bibname}{References}
\renewcommand{\bibsection}{\subsubsection*{\bibname}}

\usepackage{mathrsfs}
\usepackage{subcaption}
\usepackage{booktabs} % for professional tables

\usepackage{multirow}
\usepackage{colortbl}
\usepackage{xcolor}
\usepackage[nolist]{acronym}
\usepackage{hyperref}

\usepackage{hyperref}       % hyperlinks
\usepackage{url}            % simple URL typesetting
\usepackage{amsfonts}       % blackboard math symbols
\usepackage{nicefrac}       % compact symbols for 1/2, etc.
\usepackage{microtype}      % microtypography
\usepackage{wrapfig}
\usepackage{hyperref}
\usepackage{url}
\usepackage{graphicx}

\usepackage{amsmath,amsfonts,bm}

\def\eqref#1{equation~\ref{#1}}
\def\1{\bm{1}}

\def\rr{{\textnormal{r}}}

\def\rvtheta{{\mathbf{\theta}}}

\def\rvb{{\mathbf{b}}}
\def\rvc{{\mathbf{c}}}
\def\rvd{{\mathbf{d}}}

\def\rvr{{\mathbf{r}}}

\def\rvv{{\mathbf{v}}}
\def\rvw{{\mathbf{w}}}

\def\rvlambda{{\bm{\lambda}}}
\def\rvpi{{\bm{\pi}}}
\def\rvmu{{\bm{\mu}}}

\def\ervb{{\textnormal{b}}}

\def\ervv{{\textnormal{v}}}

\def\ervpi{{\pi}}

\def\rmW{{\mathbf{W}}}

\def\rmSigma{{\mathbf{\Sigma}}}
\def\rmLambda{{\mathbf{\Lambda}}}

\def\vv{{\bm{v}}}

\def\evc{{c}}
\def\evd{{d}}

\DeclareMathAlphabet{\mathsfit}{\encodingdefault}{\sfdefault}{m}{sl}
\SetMathAlphabet{\mathsfit}{bold}{\encodingdefault}{\sfdefault}{bx}{n}

\newcommand{\E}{\mathbb{E}}

\newcommand{\R}{\mathbb{R}}

\DeclareMathOperator*{\argmin}{arg\,min}

\newcommand{\beq}{\vspace{0mm}\begin{equation}}
\newcommand{\eeq}{\vspace{0mm}\end{equation}}

\newtheorem{lemma}{Lemma}
\newtheorem{assumption}{Assumption}

\newcommand{\kumar}{\mathrm{Kumar}}
\newcommand{\kl}{\mathrm{KL}}

\newcommand{\ie}[0]{\emph{i.e., }}

\newcommand{\eg}[0]{\emph{e.g., }}

\newcommand{\cD}{\mathcal{D}}

\begin{document}

\runningtitle{Continual Learning using a Bayesian Nonparametric Dictionary of Weight Factors}

\runningauthor{Nikhil Mehta, Kevin J Liang, Vinay K Verma and Lawrence Carin}

\twocolumn[

\aistatstitle{Continual Learning using a Bayesian Nonparametric\\Dictionary of Weight Factors}

\aistatsauthor{Nikhil Mehta$^{1}$, Kevin J Liang$^{1,2}$, Vinay K Verma$^1$, Lawrence Carin$^1$}

\aistatsaddress{$^1$Duke University \qquad $^2$Facebook AI} ]

\graphicspath{{Figures/}}

\begin{abstract}
Naively trained neural networks tend to experience catastrophic forgetting in sequential task settings, where data from previous tasks are unavailable.
A number of methods, using various model expansion strategies, have been proposed recently as possible solutions.
However, determining how much to expand the model is left to the practitioner, and often a constant schedule is chosen for simplicity, regardless of how complex the incoming task is.
Instead, we propose a principled Bayesian nonparametric approach based on the Indian Buffet Process (IBP) prior, letting the data determine how much to expand the model complexity.
We pair this with a factorization of the neural network's weight matrices.
Such an approach allows the number of factors of each weight matrix to scale with the complexity of the task, while the IBP prior encourages sparse weight factor selection and factor reuse, promoting positive knowledge transfer between tasks.
We demonstrate the effectiveness of our method on a number of continual learning benchmarks and analyze how weight factors are allocated and reused throughout the training. 
\end{abstract}

\section{Introduction}
Deep learning, trained primarily on a single task under the assumption of independent and identically distributed (\textit{i.i.d.}) data, has made enormous progress in recent years.
However, when naively trained sequentially on multiple tasks, without revisiting previous tasks, neural networks are known to suffer catastrophic forgetting~\citep{McCloskey1989, Ratcliff1990}: the ability to perform old tasks is often lost while learning new ones.
In contrast, biological life is capable of learning many tasks throughout a lifetime from decidedly non-\textit{i.i.d.} experiences, acquiring new skills and reusing old ones to learn fresh abilities, all while retaining important previous knowledge.
As we strive to make artificial systems increasingly more intelligent, natural life's ability to learn continually is an important capability to emulate.

Continual learning~\citep{Parisi2019} has attracted considerable attention recently in machine learning research, and a number of desiderata have emerged.
Models should be able to learn multiple tasks sequentially, with the eventual number and complexity of tasks unknown.
Importantly, new tasks should be learned without catastrophically forgetting previous ones, ideally without having to keep any data from previous tasks to re-train on.
Models should also be capable of positive transfer: previously learned tasks should help with the learning of new tasks.
Knowledge transfer between tasks maximizes sample efficiency, with this particularly important when data are scarce.

A number of methods~\citep{Rusu2016, Zhang2019, Lee2020} address continual learning through expansion: the model is grown with each additional task.
By diverting learning to new network components for each task, these approaches mitigate catastrophic forgetting by design, as previously learned parameters are left undisturbed.
A key challenge for these strategies is deciding when and how much to expand the network.
While it is typically claimed that this can be tailored to the incoming task, doing so requires human estimation of how much expansion is needed, which is not a straightforward process.
Instead, a preset, constant expansion is commonly employed for each new task.

\begin{figure*}[!t]
	\centering
	\includegraphics[width=0.7\textwidth]{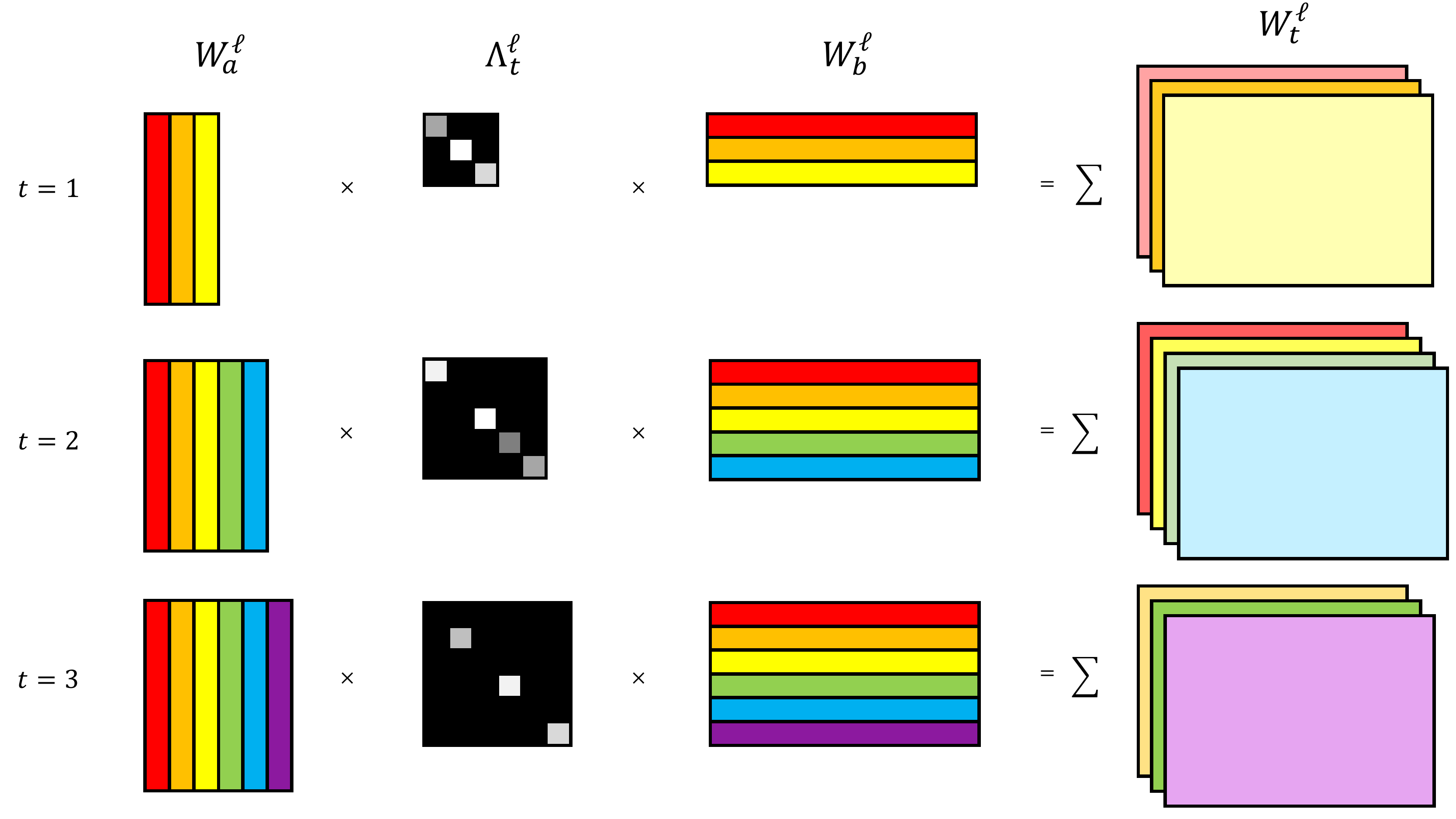}
	\caption{Layer-wise weight factors for continual learning.
	Dictionaries of weight factors $\rmW_a^\ell$ and $\rmW_b^\ell$ are shared across all tasks, and a task-specific sparse diagonal matrix $\rmLambda_t^\ell$ specifies the active factors (in this figure, elements of $\rmLambda_t^\ell$ that are black are zero, and brighter shades correspond to larger numbers).
	The weighted sum of the active weight factors yields the weight matrix for a particular task.
	The number of factors (columns of $\rmW_a^\ell$ and rows of $\rmW_b^\ell$) grows as needed with more tasks, with select factors being reused in future tasks. Best viewed in color.}
	\label{fig:IBP_W_fac}
\end{figure*}

Rather than relying on engineered heuristics, we choose to let the data dictate the model-expansion rate, employing a Bayesian nonparametric approach. 
Specifically, we couple rank-1 weight factor (WF) dictionary learning with the \ac{IBP}~\citep{Ghahramani2006}, creating a framework we call \textbf{IBP-WF}.
An IBP-based formulation allows automatic scaling of the network, but only as needed, even if the number or complexity of future tasks is unknown initially.
An IBP prior also naturally encourages recycling of previously learned skills, enabling positive transfer between tasks, which other expansion methods tend to either ignore or deal with in a more {\em ad hoc} manner.
Finally, Bayesian modeling enables model sampling, allowing for both ensembling models for increased accuracy and uncertainty estimation, which are important but rarely discussed topics in continual learning.

Our main contributions are as follows.
($i$) We introduce learning a rank-1 weight factor dictionary for a neural network expected to perform multiple tasks. We then introduce the Indian Buffet Process as a prior for each task's weight factor selection, showing why the IBP is a natural choice given continual learning's desiderata.
($ii$) We introduce a simple-but-effective method based on feature statistics for inferring task identity (ID) in incremental class settings.
($iii$) The effectiveness of IBP-WF is demonstrated on a number of continual learning tasks, outperforming other methods. We also visualize the weight factor usage across tasks, confirming both sparsity and reuse of these factors.

\section{Methods}
\subsection{Weight Factor Dictionary Learning}
\vspace{-2mm}
Consider a \ac{MLP} with layers $\ell = 1, ..., L$.
In a continual learning setting, we would like this neural network to learn multiple tasks.
Given differences between tasks, the neural network may require a different set of weight matrices $\{\rmW_t^\ell\}_{\ell=1}^{L}$ for each task $t$.
While $\{\rmW_t^\ell\}_{\ell=1}^{L}$ could be learned separately for each task, such a model does not incorporate knowledge reuse, and the total number of model parameters grows linearly with the number of tasks $T$.
While immune from catastrophic forgetting, such an approach is inefficient in both computation and data.

Instead of completely independent models for each task, we propose constituting $\rmW_t^{\ell}$ as follows:
\begin{align}
    \rmW_t^{\ell} = \rmW_a^{\ell} \rmLambda_t^\ell \rmW_b^\ell \hfill \qquad \text{s.t. } \enspace \rmLambda_t^\ell = \mathrm{diag}(\rvlambda_t^\ell) \label{eq:W_fac}
\end{align}
where $\rmW_a^{\ell} \in \R^{J \times F}$ and $\rmW_b^{\ell} \in \R^{F \times M}$ are global parameters shared across tasks and $\rvlambda_t^\ell \in \R^{F}$ is a task-specific vector. 
The determination of $F$ is explained in Section \ref{sec:IBP}, but in general $F$ is chosen such that after $T$ tasks, the total number of parameters of this factorized model is $(J+T+M) \cdot F$, which is significantly less than the $J \cdot M \cdot T$ parameters that would result from learning each task independently. We may equivalently express (\ref{eq:W_fac}) as the weighted sum of rank-1 matrices formed from the outer product of the vectors corresponding to the columns $\rmW^\ell_a$ and rows of $\rmW^\ell_b$: 
\begin{align}
    \rmW_t^{\ell} &= \sum_{k=1}^F \rvlambda_{t,k}^\ell \left(\rvw_{a,k}^{\ell} \otimes {\rvw_{b,k}^\ell}\right) \label{eq:W_fac_sum}
\end{align}
where $\otimes$ denotes the outer product, and the pair $\rvw_{a,k}^{\ell}$ and ${\rvw_{b,k}^\ell}$ is the $k^{\text{th}}$ column and row of $\rmW_{a}^\ell$ and $\rmW_{b}^\ell$, respectively.
Under this construction, the pairs of corresponding columns of $\rmW_a^{\ell}$ and rows of $\rmW_b^{\ell}$ can be interpreted as a dictionary of weight \textit{factors}, while the values in $\rvlambda_t^\ell$ are the factor \textit{scores} for a particular task (see Figure \ref{fig:IBP_W_fac}).
By sharing these global weight factors, the model can reuse features and transfer knowledge between tasks, with $\rvlambda_t^\ell$ selecting and weighting the factors for a particular task $t$.
We construct the factor scores $\rvlambda_t^\ell$ as the following element-wise product:
\begin{equation}
    \rvlambda_t^\ell = \rvr_t^\ell \odot \rvb_t^\ell
\end{equation}
where $\rvb_t^\ell \in \{0,1\}^F$ indicates the \textit{active} factors for task $t$ and $\rvr_t^\ell \in \mathbb R^F$ specifies the corresponding factor \textit{strength}. By imposing sparsity with $\rvb_t^\ell$, we concentrate skills for each task into specific factors, leaving room for learning other tasks with other factors.

\textbf{Generalizing to convolutional kernels } While learning the weight factors in (\ref{eq:W_fac}) was formulated for fully connected layers, it can be generalized to other types of layers as well, including 2D convolutional layers.
Unlike the 2D weight matrices comprising the fully connected layers of a \ac{MLP}, convolutional kernels are 4D: in addition to number of input and output channels ($C_{in}$ and $C_{out}$), they also have two spatial dimensions denoting the height ($H$) and width ($W$) of the convolutional filter.
While learning 4D tensor factors is certainly possible, in practice $H$ and $W$ tend to be small (\textit{e.g.} $H=W=3$), so we instead choose to reshape the kernel ($\R^{H \times W \times C_{in} \times C_{out}}$) into a 2D matrix ($\R^{(HWC_{in}) \times C_{out}}$). 
We then proceed with the same weight factor learning as in (\ref{eq:W_fac}). 

\subsection{Indian Buffet Process for Weight Factors}\label{sec:IBP}

Critical to the proposed layer-wise weight factors is the number of factors $F$: too few and the model lacks sufficient expressivity to model every task; too many and the model consumes more memory and computation than necessary.
To further complicate matters, the number of necessary factors likely increases monotonically as the model encounters more tasks.
While a particular choice of $F$ may be appropriate for $T$ tasks, it may no longer be sufficient after $T'$, with $T'>T$.

Rather than setting it as a constant, we let $F$ grow naturally with the number of tasks.
There are a number of expansion strategies for continual learning that have been proposed over the years~\citep{Rusu2016, Hung2019, Zhang2019, Lee2020}.
Many of these expand the model by a constant amount per task, or rely on the model designer to specify a schedule or heuristics for the size of the expansion.
These hand-tuned strategies can be brittle, and require expert knowledge on the complexity of incoming tasks.
Additionally, prior works do not use weight factor dictionaries, so expansion involves adding additional nodes to each hidden layer or learning entirely new models, which can increase test-time computation.

Instead, we employ Bayesian nonparametrics, inferring in a principled manner the scores for the proposed rank-1 weight factors for each task and the total number of factors $F$ needed.
In particular, we impose the stick-breaking construction of the \ac{IBP}~\citep{teh2007stick} as a prior for factor selection:
\begin{align}
    v^\ell_{t,i} &\sim \mathrm{Beta}(\alpha, 1) \\
    \pi^\ell_{t,k} &= \prod^{k}_{i=1} v^\ell_{t,i} \label{eq:prior_pi}\\
    b^\ell_{t,k} &\sim \mathrm{Bernoulli}(\pi^\ell_{t,k}) \label{eq:prior_b}
\end{align}
where $\alpha$ is a hyperparameter controlling the expected number of nonzero factor scores, and $k = 1, 2,...F$ indexes the factor. For the global parameters ($\rmW^\ell_a$ and $\rmW^\ell_b$) and the local factor strength ($\rvr^\ell_t$), we use point estimates. 
See Appendix~\ref{apx:review_ibp} for an overview of the \ac{IBP} and the connection of IBP-WF with IBP-based dictionary learning. Leveraging the \ac{IBP} in conjunction with dictionary learning provides a number of natural advantages within the context of continual learning:

\textbf{Dynamic control of $F$ } The \ac{IBP} allows the number of factors $F$ to be determined nonparametrically and dynamically, growing only as necessary given the complexity of each individual task. 
Simpler tasks (or ones similar to previous tasks) may require learning fewer new factors, while more complex ones lead to more, all inferred automatically.
While $F$ can theoretically grow unbounded, it does so harmonically -- much slower than the requisite linear growth of the number of tasks.

\textbf{Factor reuse and positive transfer } Given that continual learning is often deployed when tasks are at least somewhat correlated, training independent models can lead to learning redundant features, which is inefficient both in training data and test time computation.
On the other hand, the construction of $\rvpi_t^\ell$ (see (\ref{eq:prior_pi})) actively encourages reuse of existing weight factors, prioritizing recycling previously learned skills for new tasks over creating new ones, which leads to positive forward transfer.
This is in contrast to other methods whose only source of transfer is initializing from the previous task's weights, whose transfer advantage may be quickly wiped away by gradient descent.

\textbf{Catastrophic forgetting mitigation } The newly learned weight factors in $W_a$ and $W_b$ are frozen at the end of a task.
This mirrors the freezing of previously learned weights in existing expansion methods.
By blocking the gradients to weights learned from previous tasks, we avoid forgetting the model's ability to perform older tasks.
Note that while factors learned from a previous task are frozen, the factor scores may change with each incoming task allowing the model to control the usage of a previously trained factor. 

\textbf{Constant inference time cost } At test time, \ac{IBP} weight factors (outer product of each column of $\rmW_a^{\ell}$ and row of $\rmW_b^{\ell}$) can be pre-computed; given a task, the appropriate factors can be retrieved, weighted, and summed as needed to retrieve $\{\rmW_t^\ell\}_{\ell=1}^{L}$. 
Imposing the IBP prior on the usage of factors induces a prior distribution of $\mathrm{Poisson}(\alpha)$ on the number of active factors. 
Thus, the number of nonzero factors have a prior expectation of $\alpha$, regardless of the number of tasks $T$, so the expected forward computation of the model does not grow with $T$. This avoids one of the pitfalls of other expansion methods (\eg \cite{Rusu2016, Lee2020, Kumar2019}), whose inference-time computation scale with $T$.

\subsection{Variational Inference}
To determine which factors should be active for task $t$, we perform variational inference to infer the posterior of parameters $\rvtheta_t = \{\rvb^\ell_t, \rvv^\ell_t \}_{\ell = 1}^L$. We assume the following variational distributions:

\begin{align}
    q(\rvtheta^\ell_t) &= q(\rvb_t^\ell) q(\rvv_t^\ell) \label{eq:post_q}\\
    \rvb_{t}^\ell &\sim \mathrm{Bernoulli}(\rvpi^\ell_t) \label{eq:post_b}\\
    \rvv_{t}^\ell &\sim \mathrm{Kumaraswamy}(\rvc^\ell_t, \rvd^\ell_t)
\end{align}

We learn the variational parameters $\{\rvpi_t^\ell, \rvc_t^\ell, \rvd^\ell_t\}_{\ell=1}^{L}$ with Bayes by Backprop~\citep{blundell2015}.
As the Beta distribution lacks a differentiable parameterization, we use the similar Kumaraswamy distribution~\citep{Kumaraswamy1980} as the variational distribution for $\rvv_t^\ell$. 
We also use a soft relaxation of the Bernoulli distribution~\citep{MaddisonMT16} in (\ref{eq:prior_b}) and (\ref{eq:post_b}) to allow backpropagation through discrete random variables. The objective for each task is to maximize the following variational lower bound:
\beq
{
\begin{aligned}
    \mathcal L_t =& \sum_{n=1}^{N_t} \E_{q} \log{ p\left(y_t^{(n)} \big|\rvtheta_t, x_t^{(n)}, \rmW_a, \rmW_b, \rvr_t\right)}\\
    &- \underbrace{\mathrm{KL}\left( q\left(\rvtheta_t\right)||p\left( \rvtheta_t \right)\right)}_{\mathscr{R}} \hspace{-1em} \label{eq:ELBO}
\end{aligned}
}
\eeq
where $N_t$ is the number of training examples in task $t$, $\rmW_a = \{\rmW^\ell_a\}_{\ell=1}^L$, $\rmW_b = \{\rmW^\ell_b\}_{\ell=1}^L$ and $\rvr_t = \{\rvr_t^\ell\}_{\ell=1}^L$. The variational lower bound in (\ref{eq:ELBO}) is maximized with respect to task-specific parameters ($\theta_t, \rr_t$) and global parameters $(\rmW_a, \rmW_b)$. Note that in (\ref{eq:ELBO}) the first term provides label supervision and the second term ($\mathscr{R}$) regularizes the posterior not to stray too far from the IBP prior. 
We use mean-field approximation and MC sampling to approximate the second term:
\beq
{\small
\begin{aligned}
    \mathscr{R}=&  \sum_{\ell=1}^L
    \E_{q(\rvv^\ell_t)} \left[ \mathrm{KL}\left( q(\rvb^\ell_t) || p(\rvb^\ell_t|\rvv^\ell_t) \right)\right] + \mathrm{KL}\left( q(\rvv^\ell_t) || p(\rvv^\ell_t) \right) \label{eq:online_inference}
\end{aligned}
}
\eeq

where we take samples $\rvv_{t}^\ell \sim q(\rvv_{t}^\ell)$ to approximate $\E_q[\log{(p(\rvb^\ell_t|\rvv^\ell_t))}]$ in the first term. For the second term in (\ref{eq:online_inference}), we derive an analytic approximation of Kullback-Leibler (KL) divergence between two Kumaraswamy distributions. The derivation and more details on doing online inference with (\ref{eq:ELBO}) and (\ref{eq:online_inference}) are included in Appendices \ref{apx:kl} and \ref{apx:inf}.

Variational continual learning (VCL)~\citep{Nguyen2018} addresses catastrophic forgetting using online inference, $i.e.$, the posterior inferred from the most recent task is used as a prior for the incoming task. However, recent work~\citep{farquhar2018robust, farquhar2019unifying} suggests that approximate online inference often does not succeed in mitigating catastrophic forgetting in realistic continual learning settings, as methods based solely on approximate inference rely on a simple prior to capture everything learned on all previous tasks. Thus, instead of performing online inference for all parameters $\{\rvr^\ell_t, \rvb_t^\ell, \rvv^\ell_t\}$, we only apply online inference for $\rvv^\ell_t$ and learn task-specific parameters $\{\rvr^\ell_t, \rvb_t^\ell\}$. Note that online inference over $\rvv^\ell_t$ encourages the reuse of factors from previous tasks while having task-specific parameters allows the model to easily adapt to a new task by using new factors.

\textbf{Preserving knowledge } If all of $\rmW_a^{\ell}$ and $\rmW_b^{\ell}$ were free to move without constraint, then catastrophic forgetting may still occur.
Indeed, the model could ``reuse'' a factor from a previous task and then repurpose it entirely, undermining the ability to do the former task.
To prevent this, the weight factors (\ie the columns of $\rmW_a^{\ell}$ and rows of $\rmW_b^{\ell}$) with factor probability $\pi^\ell_{t,k} > \kappa$ are locked (\textit{e.g.}, with a stop gradient operator) at the conclusion of a task.
Weight factors below the threshold $\kappa$ are left free to be modified by future tasks.
Throughout our experiments, we set the threshold as $\kappa=0.5$, but this can be adjusted based on tolerance for forgetting. We include an ablation study on selecting $\kappa$ in the Appendix~\ref{apx:ablation_kappa}.
Alternatively, other regularization methods (\eg \cite{Kirkpatrick2017}) can be used to prevent important factors from drifting too far, but we leave this combination to future work.

\subsection{Task Inference at Test Time}
\label{sec:task_infer}
IBP-WF addresses catastrophic forgetting and allows for positive knowledge transfer.
However, as with many continual learning methods, IBP-WF requires the task identity associated with each input at test time in order to select the proper $\{\rmLambda_t^\ell\}_{\ell=1}^{L}$.
The validity of this assumption has occasionally been questioned~\citep{farquhar2018robust, Aljundi2019b, Lee2020}.
We outline here a mechanism for enabling IBP-WF to operate in an incremental class setting, inferring the task identity at test time. 
Given a data point $x$, we can infer the task identity by defining the probability of $x$ belonging to a particular task $t$ as follows:
\beq
{
\begin{aligned}
    P(t|x) &\propto P(x|t)P(t) \label{eq:gen_task_inference}
\end{aligned}
}
\eeq
However, using (\ref{eq:gen_task_inference}) requires learning a generative model $P(x|t)$ $\forall t \in \{1,2,...,T\}$, which can be expensive in both computation and the number of parameters. 
To alleviate this issue, we propose a simple yet effective alternative: we define an approximation to $P(x|t)$ by using the feature distribution induced by an intermediate hidden layer of the trained neural network. 
In particular, we approximate (\ref{eq:gen_task_inference}) by using $P(t|\phi(x))$ as a surrogate for $P(t|x)$:
\beq
{
\begin{aligned}
    P(t|x) &\approx P(t|\phi(x)) = \frac{P(\phi(x)|t)P(t)}{\sum_{t'} P(\phi(x)|t')P(t')} \label{eq:phi_task_inference}
\end{aligned}
}
\eeq
where $\phi$ is an intermediate layer defined using the proposed weight factorization as shown in (\ref{eq:W_fac}) with task-specific weights of the first task. 
We work with the feature space induced by the parameters of the first task as they are accessible by the training data of all tasks that follow. 
Next, we assume $P(\phi(x)|t)$ to be a Gaussian distribution: $P(\phi(x)|t) = \mathcal{N}(\phi(x)|\mu_t, \Sigma_t)$, where the parameters are the empirical estimates using the training data.

\beq
{
\begin{aligned}
\hat{\mu}_{t} &= \frac{1}{N_t} \sum_{n=1}^{N_t} \phi(x^{(n)}_{t}), \\ \hat{\Sigma}_{t} &= \frac{1}{N_t} \sum_{n=1}^{N_t} (\phi(x_t^{(n)})-\hat{\mu}_t)(\phi(x_t^{(n)})-\hat{\mu}_t)^T \label{eq:parameter_est}
\end{aligned}
}
\eeq
where $x_t^{(n)}$ is a training sample from task $t$. 
When we train our model on task $t>1$, we use the task-specific weights learned for the first task to compute $\{\rvmu_t, \rmSigma_t\}$.  
The parameters $\{\mu_t, \Sigma_t\}$ are stored to infer test-time task identity. 
While the features may not be exactly Gaussian distributed, this assumption has been shown to work well in deep learning \citep{heusel2017gans, LeeNIPS2018}, and we find it effective in practice; see Appendix~\ref{apx:task_results} for task inference accuracy experiments and t-SNE visualizations of $\phi$.
Notably, we achieve similar accuracy to generative model task inference~\citep{Lee2020}, with a far cheaper method.

Considering the marginal task distribution $P(t) \propto N_t$, the task identity can be inferred as follows:
\beq
{
\begin{aligned}
    \hat{t} = \argmin_t \Bigg[& \frac{\log{|\hat{\Sigma}_t|}}{2}
    - \log{(N_t)}  \\
    & + \frac{1}{2}(\phi(x) - \hat{\mu}_t)^T \hat{\Sigma}_t^{-1}(\phi(x) - \hat{\mu}_t) \Bigg] \label{eq:task_inf}
\end{aligned}
}
\eeq
where $\hat{t}$ is the inferred task and $I$ is the identity matrix. 
While such a strategy does require storing statistics $\hat{\mu}_t$ and $\hat{\Sigma}_t$, the total size of these is still considerably smaller than parameter statistics required by certain regularization methods (\eg EWC~\citep{Kirkpatrick2017}), as well as the coresets or generative models used by replay methods~\citep{Nguyen2018, Shin2017, Van2018}.

\textbf{Remark:} (Informal) The approximation in (\ref{eq:phi_task_inference}) is exact if $\phi$ is an invertible map since $P(s|t) = M \times P(x|t)$ with $M = \left| \mathrm{det} \frac{ \partial \phi^{-1}}{\partial s} \right|$ when $s = \phi(x)$. 
See Appendix \ref{apx:task_infer_proof} for a formal proof.

\section{Related Works}

There have been a number of diverse continual learning methods that have been proposed in recent years, most of which can be roughly grouped by strategy into a few categories, with some overlap.
Regularization-based approaches~\citep{Kirkpatrick2017, Zenke2017, Li2017, Nguyen2018, Aljundi2018, Schwarz2018, Ritter2018} add a loss term constraining the network parameters to remain close to solutions of previously learned tasks.
Others use replay~\citep{Kirkpatrick2017, Lopez2017, Shin2017, Nguyen2018, Rolnick2019}, which retrains the model on samples from earlier tasks, either from a saved core set or with a generative model that must be learned.

Another class of continual learning methods rely on expansion, the approach taken by IBP-WF. 
Progressive Neural Networks~\citep{Rusu2016} learn a new neural network column for each new task, with previous columns' features as additional inputs.
While avoiding catastrophic forgetting by design, both memory and computation grow linearly with the number of tasks $T$, just as if one were to learn independent models per task.
Side-tuning \citep{Zhang2019} learns a separate ``side'' network for each task, adding the output to a shared base model; while this experiences linear growth $T$ of the model size, it reduces the cost by keeping each side network small.
As an alternative to constant growth, Reinforced Continual Learning~\citep{xu2018reinforced} uses an LSTM~\citep{Hochreiter1997lstm} controller and REINFORCE~\citep{williams1992simple} to determine the expansion rate, while Dynamically Expandable Networks~\citep{yoon2017lifelong} expand by a constant amount before using sparsity regularization and loss-based heuristics to prune away unused units.
Pruning between tasks is also utilized by \citet{Hung2019}, where the pruning and re-training is used to prevent excessive growth of the model.
MNDPT~\citep{veniat2021efficient} adds new modules to the model with new tasks, reusing modules of older similar tasks.

\begin{table*}[!t]
\centering
\caption{\label{tbl:mnist_exp}
The average accuracy of seen tasks after learning on a sequence of tasks using a \ac{MLP}.}
\resizebox{\textwidth}{!}{%
\begin{tabular}{lccccc}
\toprule
Method & Replay & \multicolumn{2}{c}{Split MNIST} & \multicolumn{2}{c}{Permuted MNIST} \\ \midrule
& & Incremental Task & Incremental Class & Incremental Task & Incremental Class \\ \toprule
% SGD  &      & 97.98 $\pm$ 0.09 & 19.46 $\pm$ 0.04 & 94.94 $\pm$ 0.24 & 14.02 $\pm$ 1.25 \\
% Adam &      & 93.46 $\pm$ 2.01 & 19.71 $\pm$ 0.08 & 93.42 $\pm$ 0.56 & 12.82 $\pm$ 0.95\\
Adagrad     & & 98.24 $\pm$ 0.59 & 19.73 $\pm$ 0.12 & 90.78 $\pm$ 0.18 & 27.59 $\pm$ 1.07\\ 
\arrayrulecolor{gray}\cmidrule{1-6}\arrayrulecolor{black}
% $L_2$      & & 98.18 $\pm$ 0.96 & 22.52 $\pm$ 1.08 & 95.45 $\pm$ 0.44 & 13.92 $\pm$ 1.79 \\ 
EWC & & 98.64 $\pm$ 0.87 & 19.89 $\pm$ 0.04 & 92.49 $\pm$ 0.34 & 23.97 $\pm$ 3.21 \\
% Online EWC              & & 98.04 $\pm$ 1.10 & 19.77 $\pm$ 0.04 & 95.15 $\pm$ 0.49 & 42.58 $\pm$ 6.50 \\ 
SI & & 99.16 $\pm$ 0.52 & 19.71 $\pm$ 0.10 & 95.45 $\pm$ 0.59 & 56.88 $\pm$ 4.93         \\
MAS  & & 99.23 $\pm$ 0.18 & 19.58 $\pm$ 0.11 & 96.76 $\pm$ 0.26 & 49.95 $\pm$ 2.53     \\
LwF & & 99.61 $\pm$ 0.05 & 22.31 $\pm$ 0.51 & 81.47 $\pm$ 0.38 & 30.63 $\pm$ 0.76       \\
VCL & & 96.79 $\pm$ 0.35 & 19.43 $\pm$ 0.02 & 91.33 $\pm$ 0.93 & 16.21 $\pm$ 0.59\\
\arrayrulecolor{gray}\cmidrule{1-6}\arrayrulecolor{black}
Naive rehearsal         & \checkmark & 99.39 $\pm$ 0.11 & 85.97 $\pm$ 0.75 & 96.75 $\pm$ 0.19 & 96.53 $\pm$ 0.11 \\
VCL-Coreset & \checkmark & 98.75 $\pm$ 0.06 & 85.15 $\pm$ 0.61 & 93.46 $\pm$ 0.49 & 66.96 $\pm$ 4.10\\
GEM & \checkmark & 98.56 $\pm$ 0.08 & 88.28 $\pm$ 0.26 & 97.14 $\pm$ 0.09 & 96.88 $\pm$ 0.05 \\
DGR                   & \checkmark & 99.54 $\pm$ 0.05 & 91.61 $\pm$ 0.26 & 93.74 $\pm$ 0.24 & 92.96 $\pm$ 0.53 \\
RtF                     & \checkmark & 99.66 $\pm$ 0.03 & \textbf{92.56} $\pm$ 0.21 & 97.31 $\pm$ 0.01 & 96.23 $\pm$ 0.04\\
\midrule 
\textbf{IBP-WF} (Ours) & & \textbf{99.69} $\pm$ 0.05 & 92.40 $\pm$ 0.15 & \textbf{97.52} $\pm$ 0.06 & \textbf{97.50} $\pm$ 0.06 \\ 
\bottomrule
\end{tabular}
}
\end{table*}

A few works have also explored continual learning from a Bayesian nonparametric perspective. \cite{Lee2020} combine the Dirichlet process with a mixture of experts, where each expert is a neural network responsible for a subset of the data. While this approach does allow the data to dictate model expansion, mixing only occurs at the prediction representation, as opposed to throughout the model as in IBP-WF. This mixture of experts thus can lead to redundant feature learning and unnecessary extra computation. 
Recently, there have been other attempts to apply \ac{IBP} to learn the structure of a neural network for continual learning. \cite{Kumar2019} proposed Bayesian Structure Adaptation for Continual Learning (BSCL), which expands the hidden units in each layer using a binary mask for the weight filters, with an \ac{IBP} as the prior of the mask. Since BSCL uses an $\ac{IBP}$ over the entire weight matrix, the inference parameters grow quadratically with the layer size requiring more memory and making it hard to scale to large networks, whereas we use the $\ac{IBP}$ to model the factor scores where the inference parameters only grow linearly with the layer size. H-\ac{IBP} Bayesian Neural Networks (HIBNN) \citep{kessler2020hierarchical} uses sequential Bayes to apply hierarchical \ac{IBP} to the hidden layer activations of a fully connected neural network. In contrast, we expand the number of factors of the weight matrix in each layer, allowing us to scale our method to deeper networks with convolutional layers.

\section{Experiments}
\label{sec:experiments}
We evaluate our method in two settings, which we call incremental \textit{task} learning and incremental \textit{class} learning. 
In incremental task learning, the task identity (ID) of each sample is revealed at test time.
In this case, we can simply use the $\Lambda_t$ from the task ID given. 
On the other hand, in incremental class learning, we are not given task IDs during testing.
This is the more difficult case, with many earlier continual learning methods tending to do poorly.
We address this challenge by using the approach described in Section~\ref{sec:task_infer}, inferring the task identity by using the training statistics at an intermediate layer. 
For task inference in our incremental-class experiments, we consider $\phi$ in (\ref{eq:phi_task_inference}) to be the representation after the first layer, as it performed best. {We purposely have chosen to {\em not} supplement IBP-WF with replay, to isolate the advantages of using IBP and weight factors. This puts IBP-WF at a disadvantage compared to replay-based methods. Nevertheless, IBP-WF outperforms or is comparable to replay-based methods. Note that {\em with} replay (where we no longer freeze the IBP ``dishes'' after they are learned by a given task), the IBP-WF performance is likely to improve further (via backward transfer); we reserve that for future work.}

We additionally perform an ablation study over the effect of the IBP, and then visualize some IBP-WF weight factors to verify some of its behavior. The description of baselines and the training details are in Appendices~\ref{apx:baselines} and \ref{apx:setup}, respectively.
Ablation studies for IBP-WF's $\alpha$ and $\kappa$ are also included in Appendices~\ref{apx:ablation_alpha} and~\ref{apx:ablation_kappa}.
All experiments are run on a NVIDIA Titan X GPU.
\begin{figure*}[!t]
    \centering
    \includegraphics[width=\textwidth]{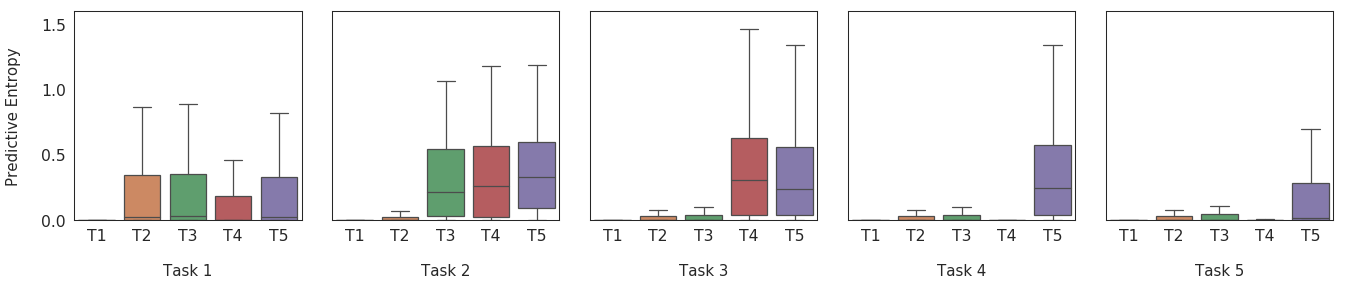}
    \caption{Uncertainty in the incremental class setting for Split MNIST dataset. Each plot depicts the uncertainty on the test sets after training on each task sequentially. The $y$-axis denotes the uncertainty (as the predictive entropy in nats), and $x$-axis denotes the test sets ($\mathcal{T}_1$ through $\mathcal{T}_5$) for each task.}
    \label{fig:uncertainty_incr_class}
\end{figure*}
\begin{figure*}[!ht]
\begin{subfigure}{.49\textwidth}
  \includegraphics[width=\linewidth]{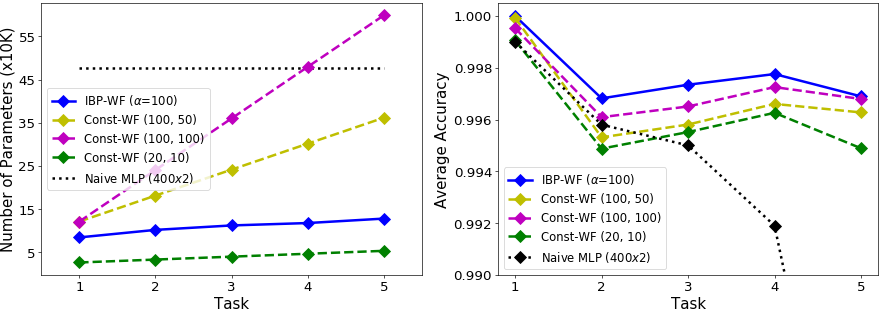}  
  \caption{Split MNIST (5 tasks)}
  \label{fig:splitmnist}
\end{subfigure}\hspace{0.02\textwidth}
\begin{subfigure}{.49\textwidth}
  \includegraphics[width=\linewidth]{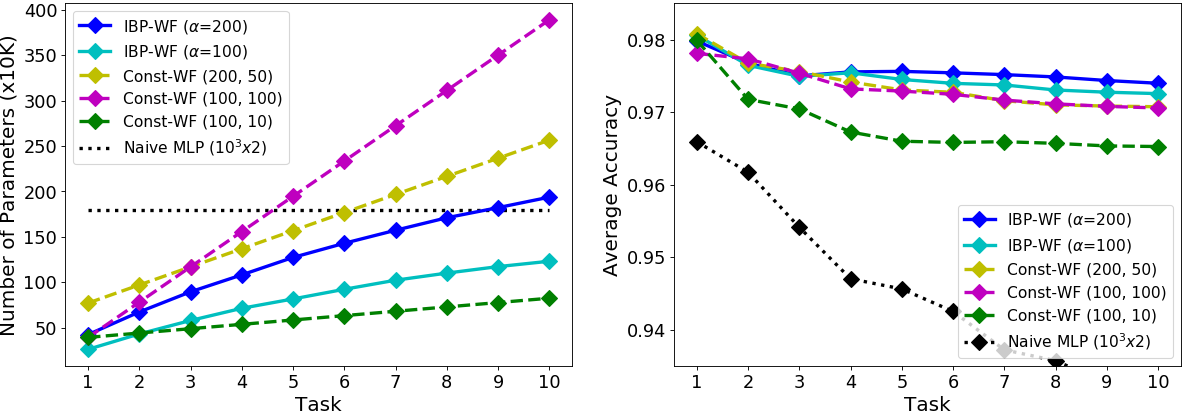}  
  \caption{Permuted MNIST (10 tasks)}
  \label{fig:permmnist}
\end{subfigure}
\caption{Parameter usage (left) and Average accuracy (right) for IBP-WF and Const-WF. We see that IBP-WF performs favorably by learning the required number of factors for each task in contrast to Const-WF.}
\label{fig:const-wf}
\end{figure*}

\subsection{Datasets and Architectures}
\label{sec:analysis}
We evaluate IBP-WF on a number of common continual learning benchmarks.
For each, the model is trained on a series of classification tasks arriving in sequence. 
This is done \textit{without} revisiting the data from previous tasks, unless otherwise stated (\eg some baselines use a memory buffer for replay to relax this constraint). 
The standard train/validation/test splits were used.

\textbf{Split MNIST} Following \cite{Zenke2017}, the 10 digit classes of the MNIST \citep{Lecun1998} dataset are split into a series of 5 binary classification tasks: 0 vs 1, 2 vs 3, 4 vs 5, 6 vs 7, and 8 vs 9. 
In the incremental task setting, where the task ID is given, this reduces to a binary classification problem during testing. 
In the incremental class setting, without task labels, each model must predict one of $2t$ classes, up to a maximum of 10 once all tasks have been seen.

\textbf{Permuted MNIST} First used to characterize catastrophic forgetting in neural networks by \cite{Goodfellow2013}, Permuted MNIST has remained a common continual learning benchmark.
The first classification task is typically chosen to be the MNIST dataset, unchanged. 
Each subsequent task consists of the same 10-way digit classification, but with the pixels of the entire MNIST dataset randomly permuted in a consistent manner.
An arbitrary number of tasks can be generated in this manner; for our experiments, we use 10 tasks. 
In the incremental-task setting, test-time evaluation is a 10-way classification problem, while in incremental class learning we have up to 100 classes.

Results are shown in Table~\ref{tbl:mnist_exp}. In addition to these results, we also compare against additional baselines in Appendix \ref{apx:baselines}.
IBP-WF outperforms other methods in most cases and is inferior only to replay-based methods in one setting. Unlike replay-based methods though, IBP-WF does not require saving data examples or separately learning bulky generative models. Compared with non-replay methods, we see significant improvement, especially in the incremental class setting. Additionally, due to the Bayesian nature of IBP-WF, one can quantify the predictive uncertainty, which is a desirable property of a model, especially in decision making. Uncertainty estimates can also be used to detect out-of-distribution samples \citep{Gal2016Uncertainty} and adversarial attacks~\citep{smith2018understanding}.
We demonstrate the former in Figure~\ref{fig:uncertainty_incr_class}, where data from unseen tasks can be identified by the model's significantly higher uncertainty.
See Appendix \ref{apx:uncertainty} for additional details on uncertainty estimation methodology.

\begin{figure*}[!ht]
    \centering
    \includegraphics[width=0.9\textwidth]{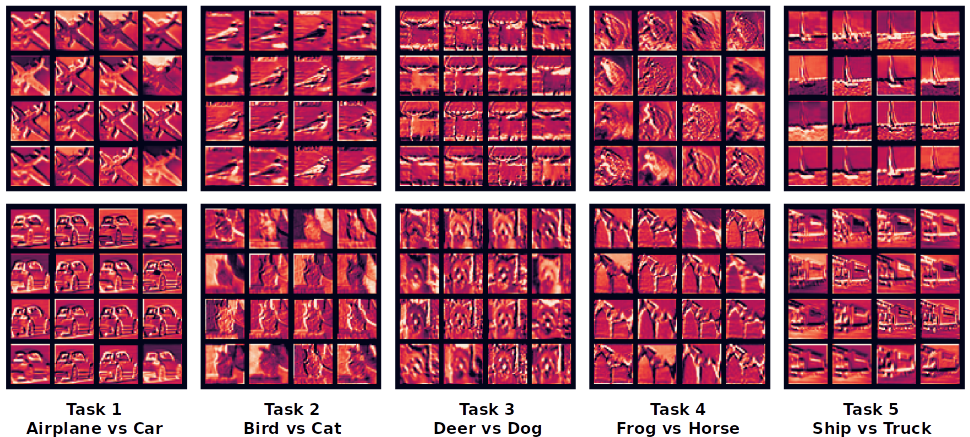}
    \caption{The first layer representations for each class in a trained IBP-WF ResNet-20. Each $4\times 4$ grid shows the feature representations after convolution using 16 kernels in the first layer.} 
    \label{fig:w_viz}
\end{figure*}

\textbf{Split CIFAR10} We split the CIFAR10~\citep{Krizhevsky2009} dataset into a sequence of 5 binary classification tasks (see Figure~\ref{fig:w_viz} for the class pairings). Similar to Split MNIST, this is a binary classification problem at test time in the incremental task setting, and $2t$-wise classification in the incremental class setting.

\begin{table}[t]
% \begin{wraptable}{r}{0.5\textwidth}
% \vspace{-3mm}
\centering
% \scriptsize
\caption{\small \label{tbl:split_cifar_exp}
The average accuracy of all seen tasks after learning the task sequence.}
\resizebox{0.49\textwidth}{!}{
\begin{tabular}{lccc}
\toprule
Method & Replay & \multicolumn{2}{c}{Split CIFAR10} \\ \midrule
& & Incremental Task & Incremental Class \\ \toprule
% SGD  & &66.60 $\pm$ 7.38 & 19.52 $\pm$ 0.03 \\
% Adam & &62.49 $\pm$ 6.01  & 19.61 $\pm$ 0.01\\
Adagrad & & 71.56 $\pm$ 1.73 & 19.59 $\pm$ 0.02\\
\arrayrulecolor{gray}\cmidrule{1-4}\arrayrulecolor{black}
$L_2$ & & 74.36 $\pm$ 0.83 & 16.86 $\pm$ 0.08\\ 
EWC & & 75.91 $\pm$ 1.64 & 18.84 $\pm$ 0.06\\
Online EWC & & 88.34 $\pm$ 1.06 & 17.54 $\pm$ 0.34\\ 
SI & & 87.19 $\pm$ 2.06 & 19.06 $\pm$ 0.09\\
MAS  & & 85.68 $\pm$ 1.36 & 16.29 $\pm$ 0.14 \\
\arrayrulecolor{gray}\cmidrule{1-4}\arrayrulecolor{black}
Naive rehearsal & \checkmark & 87.79 $\pm$ 0.88& 34.24 $\pm$ 1.38\\
\midrule 
\textbf{IBP-WF} (Ours) & & \textbf{90.94} $\pm$ 2.65 & \textbf{40.40} $\pm$ 0.21\\ \bottomrule
\end{tabular}
}
% \vspace{-4mm}
% \end{wraptable}
\end{table} 
\begin{figure*}[!t]
    \centering
    \includegraphics[width=0.95\textwidth]{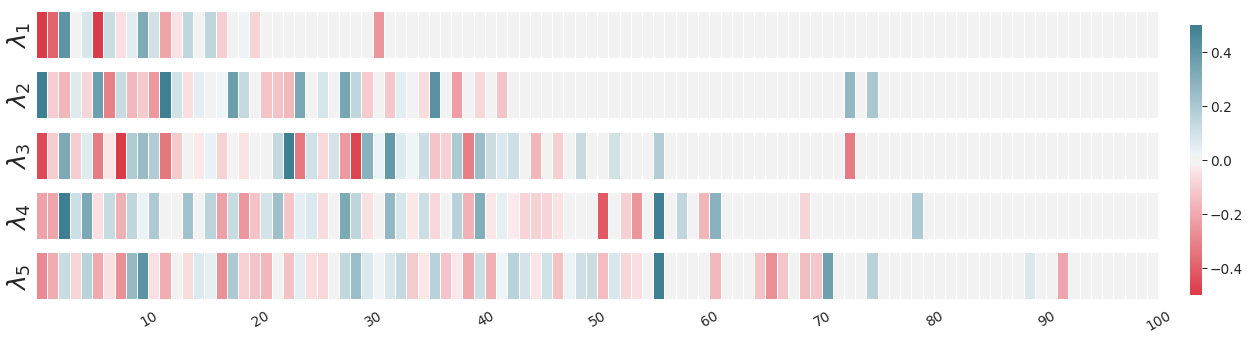}
    \caption{(Horizontal axis: Weight factors in the dictionary,  vertical axis: Factor scores for 5 tasks in Split CIFAR10). High overlap in \emph{earlier} factor scores indicate that tasks reuse factors that were learned by previously seen tasks. 
    The IBP prior encourages the reuse of previously learned factors, while inducing sparsity on the total number of active factors.}
    \label{fig:lambda}
\end{figure*}

State-of-the-art classification performance on CIFAR10 is typically achieved with convolutional neural networks. We demonstrate that IBP-WF can scale by using ResNet-20~\citep{He2016} as our architecture, with separate batch normalization layers for each task. For task inference at test time, we take the average across spatial dimensions $H$ and $W$ to get the feature statistics $\phi$ and then proceed with the parameter estimation procedure introduced in (\ref{eq:parameter_est}). We keep a buffer of 400 images from previous tasks for the naive rehearsal baseline.
Table~\ref{tbl:split_cifar_exp} shows the results on Split CIFAR10.
We again see that IBP-WF performs well relative to the baseline methods.

\subsection{IBP Ablation Study}
To demonstrate the benefits of the IBP prior, we perform an ablation study comparing IBP-WF with a variant without the IBP for weight factor selection and expansion, in which factor usage and model growth must be manually set.
As the expansion rate is constant, we call this Const-WF$(\nu, \omega)$, parameterized by the starting number of factors ($\nu$) in the first task and number of new factors ($\omega$) added per task.
We compare IBP-WF with Const-WF for several corresponding settings in Figure~\ref{fig:const-wf}.
Importantly, the IBP allows for automatic expansion as needed, with sublinear harmonic parameter growth, without human intervention. 
IBP-WF's factor reuse also results in positive transfer, yielding better accuracy, despite having fewer parameters than Const-WF. 

\subsection{Visualizations}
\textbf{Weight factor utilization} 
Central to our method is the IBP prior that controls the growth of the number of factors and encourages the model to reuse factors.
This controlled growth makes IBP-WF more efficient than independent models, while the reuse allows for positive knowledge transfer between tasks.
The factor usage is visualized by plotting the expected factor scores $\E\left[\rvlambda_t\right]$ for the first layer of a model trained on the Split CIFAR10 in Figure~\ref{fig:lambda}.
One can clearly see the impact of using the IBP as regularization: early factors are prioritized in earlier tasks, and new factors are used with later tasks.
We emphasize that the number of new parameters is not defined directly by some preset schedule, but rather is inferred from the data.

The sparsity induced by the IBP can also be seen. 
With each new task, an increasing number of factor scores have nonzero entries, as the model adapts the number of factors $F$ based on the task objective. 
However, even for a later task, the probability of a factor being active remains high for only a few.
As a result, each \textit{draw} from the posterior tends to be sparse, regularized by the IBP to have $\alpha$ active factors in expectation. Another appealing aspect of using an IBP is that the \textit{rate} of allocating a new factor decreases with tasks. Finally, following the ``rich-get-richer'' principle (here for ``rich'', or widely utilized, factors), the \ac{IBP} encourages that factors are reused based on the total number of prior tasks using it.

\textbf{CIFAR10 filters } We also visualize the first layer convolutional representations for a model trained with IBP-WF on Split CIFAR10 (Figure~\ref{fig:w_viz}). We observe an interesting property of the model: the feature maps in earlier tasks are similar compared to the diverse feature maps for later tasks. 
This can be attributed to early tasks using few factors due to the regularizing effect induced by the \ac{IBP} on the rank of the weight filter. However, as the model observes more data, the filters become more diverse since the number of active factors increase, resulting in varied features maps. This shows that as the model sees more tasks, the complexity of the layer increases through the newly invoked filters.

\section{Conclusions}
An expansion-based approach combining a dictionary of weight factors with the IBP has been introduced, which we call IBP-WF.
This synergy provides important characteristics within the context of continual learning, including knowledge reuse across tasks, data-driven model expansion, and catastrophic-forgetting mitigation.
We also propose a simple and efficient task-inference scheme, utilizing feature statistics for each task and enabling operation in incremental class settings.
A number of experiments on common continual-learning benchmarks show the effectiveness of IBP-WF. 
Ablation studies demonstrate the effectiveness of the IBP over linear expansion, and visualizations of the inferred factor scores and weights illustrate the regularization effects of our method. 
Notably, the motivation of IBP-WF is orthogonal to a number of other continual learning strategies, and combining some of these with IBP-WF is a promising direction for future work. 
For example, IBP-WF can readily be augmented with replay, and the Dirichlet process mixture model (as in \cite{Lee2020}) may be a natural Bayesian nonparametric alternative to our feature statistic method for inferring tasks.

\small
\textbf{Acknowledgements:} This work was supported by the DARPA L2M Program.
\normalsize

{
\small
\bibliography{bibliography}
\bibliographystyle{bibliography}
}

\clearpage
\onecolumn
\appendix
\aistatstitle{Supplementary Material}

\section{Review of \acf{IBP} and its connection to IBP-WF}
\label{apx:review_ibp}
\textbf{Indian Buffet Process} The \acf{IBP} is a stochastic process defining a probability distribution over a binary matrix ($Z$) with finite rows ($T$) and an unbounded number of columns ($K \rightarrow \infty$). The binary matrix can be interpreted as an assignment matrix, with the rows representing a finite number of objects (sometimes referred to as the ``customers'') and the columns representing an unbounded number of features (referred to as the ``dishes''), where $z_{ik} = 1$ if an object $i$ has the $k^{th}$ feature or otherwise $z_{ik} = 0$. Consider $z_{ik}|\mu_k \sim \text{Bernoulli}(\mu_k)$, where $\mu_k \sim \text{Beta}(\frac{\alpha}{K}, 1)$ is the prior probability that the feature $k$ is active and $\alpha$ is the strength parameter. If we marginalize $\mu_k$ and take the limit $K\rightarrow\infty$, we get the \ac{IBP}. The \ac{IBP} is often described using a culinary metaphor: supposing that there is a restaurant that serves a buffet with infinitely many dishes, then we can describe the \ac{IBP} as follows: 
\begin{itemize}
    \setlength{\itemsep}{0.5em}
    \item The first customer enters the restaurant and takes a serving from Poisson$(\alpha)$ dishes.
    \item Each $t^{\mathrm{th}}$ customer that follows moves along the buffet sampling dishes based on their popularity; the customer takes a serving of the $k^{\mathrm{th}}$ dish with the probability $\frac{m_k}{t}$, where $m_k$ is the number of customers who have previously taken dish $k$. The customer then tries Poisson$(\frac{\alpha}{t})$ number of \emph{new} dishes.
\end{itemize}
A sample from the above process can be summarized with the binary matrix $Z$, where $z_{ik}$ represents whether the $i^{\mathrm{th}}$ customer took a serving from dish $k$ or not. For an in-depth view, we refer the reader to the comprehensive review by \cite{Ghahramani2006}.

\textbf{Key properties of \ac{IBP}}: (1) The total number of dishes chosen can grow arbitrarily. (2) The likelihood of adding new dishes is given by Poisson$(\frac{\alpha}{t})$. Thus, as $t$ increases, the tendency to add new dishes decreases. (3) As the number of customers increase, the tendency of a new customer to reuse previously served dishes increases.

\textbf{Stick-breaking Construction for \ac{IBP}} \cite{teh2007stick} proposed an alternative representation for the \ac{IBP} where the feature probabilities ($\mu_k$) are not integrated. Let the ordered sequence of $\{\mu_k\}_{k=1}^K$ be $\pi_1 > \pi_2 > ... > \pi_K$ such that $\pi_k = \mu_l$, where $1 \leq \{k,l\} \leq K$. We can construct $\{\pi_k\}_{k=1}^K$ as follows: 
\begin{align}
    v_k \overset{i.i.d.}{\sim} \text{Beta}(\alpha, 1) \qquad \qquad \pi_k = \prod_{j=1}^k v_l
\end{align}
In the limit $K\rightarrow\infty$, the above is referred to as the stick-breaking construction of IBP. The stick-breaking construction and the standard \ac{IBP} representation are different representations of the same nonparametric object (see Section 3 in \citet{teh2007stick} for the proof). In practice, we use a truncated version of the stick-breaking process, where a large enough $K$ is chosen.

\textbf{Connection with the proposed IBP-WF for continual learning} Recall the filter construction in the proposed weight factor dictionary learning:
\begin{align}
    W_t = \sum_{k=1}^F \lambda_t \left( w_{a,k} \otimes w_{b,k} \right), \qquad \qquad \rvlambda_t = \rvr_t \odot \rvb_t
\end{align}
where $F$ represents the number of active factors, $t$ represents the task and $k$ represents the row and column of the corresponding $W_a$ and $W_b$ matrices. For brevity, here we suppress the superscript $\ell$ from equation~(\ref{eq:W_fac_sum}) denoting the layer. The binary vector $b_t$ is generated using the stick-breaking construction of \ac{IBP}. Following the same culinary metaphor as in the standard \ac{IBP}, the weight factor $\left( w_{a.k} \otimes w_{b,k}\right)$ and the task $t$ are analogous to the ``dish'' and ``customer''  respectively. Central to our setup is the growth of $F$ as the model encounters new tasks; IBP-WF inherits the properties of \ac{IBP} described above: as more tasks are seen, the rate of adding new weight factors decreases while the likelihood of reusing previously learned factors simultaneously increases.

\section{\ac{KL} Divergence Derivation}
\label{apx:kl}
\cite{nalisnick2016stickbreaking} gave an approximate form for the \ac{KL} divergence between Kumaraswamy and Beta distributions. We apply online variational inference for $v$, which requires the \ac{KL} divergence between two Kumaraswamy distributions (See online inference for $\vv^{\ell}_t$ in Appendix~\ref{apx:inf}). Here we derive the analytical form to approximate the \ac{KL} divergence between two Kumaraswamy distributions $q$ and $p$.  
\beq
\begin{aligned}
\kl\left( q_v\left( a, b \right) || p_v\left( \alpha, \beta \right) \right) &= \E_{q_v} \left[ \log{ \frac{q_v\left(a, b \right)}{p_v\left(\alpha, \beta \right)} } \right]
\end{aligned}
\eeq
where  $q_v\left( a, b \right) = ab v^{a-1} \left(1-v^{a}\right)^{b-1} \enspace \text{and} \enspace p_v\left( \alpha, \beta \right) = \alpha \beta v^{\alpha-1} \left(1-v^{\alpha}\right)^{\beta-1}$.

\beq
\begin{aligned}
\small
\kl\left( q_v\left( a, b \right) || p_v\left( \alpha, \beta \right) \right) =&  \underbrace{\E_{q_v} \left[ \log{ q_v\left(a, b \right) } \right]}_{\mathscr{T}1} - \underbrace{\E_{q_v} \left[ \log{p_v\left(\alpha, \beta \right)} \right]}_{\mathscr{T}2} \\
\end{aligned}
\eeq
where the first term is the Kumaraswamy entropy \citep{kumarentropy}:
\beq
\begin{aligned}
\mathscr{T}1 &=  \log{ab} + \frac{a-1}{a} \left( -\gamma -\Psi(b) - \frac{1}{b} \right) - \frac{b-1}{b} \label{eq:kl1}
\end{aligned}
\eeq
where $\gamma$ is Euler's constant and $\Psi$ is the Digamma function. For the second term, we write the expectation as:
\begin{align}
\mathscr{T}2 &= \E_{q_v} \log{ \left( \alpha \beta v^{\alpha-1}  \left(1-v^{\alpha}\right)^{\beta-1} \right) } \label{eq:kl2} \\ 
&= \E_{q_v} \left[ \log{\alpha \beta} + \left( \alpha-1 \right)\log{v} + \left( \beta-1 \right) \log{ \left( 1 - v^\alpha \right)} \right] \nonumber \\
&= \log{\alpha \beta} + \left( \alpha-1 \right) \E_{q_v} \log{v} + \left( \beta-1 \right) \E_{q_v} \log{ \left( 1 - v^\alpha \right)} \nonumber
\end{align}
In the above equation, the expectation of the $\log$ term can be computed using \cite{gradshteyn2007} (4.253):
\beq
\begin{aligned}
\E_{q_v} \log{v} &= \frac{1}{a} \left( -\gamma - \Psi(b) - \frac{1}{b} \right)
\end{aligned}
\eeq
The third term involves taking the expectation of $\log{\left( 1-v^\alpha \right)}$ which can approximated with a Taylor series: 
\beq
\begin{aligned}
\log{ \left( 1 - v^\alpha \right)} &= -\sum_{m=1}^\infty \frac{1}{m}v^{m\alpha} \label{eq:taylor_expansion}
\end{aligned}
\eeq
Note that the infinite sum in (\ref{eq:taylor_expansion}) converges since $0 < v < 1$. From the monotone convergence theorem, we can take the expectation inside the sum:
\beq
\begin{aligned}
\E_{q_v} \left[ \log{ \left( 1 - v^\alpha \right)} \right] &= -\sum_{m=1}^\infty \frac{1}{m} \E_{q_v} v^{m\alpha} \\
&= -\sum_{m=1}^\infty \frac{b}{m} \,\mathrm{B}\left(\frac{m\alpha}{a}+1, b\right)\\
&= -\sum_{m=1}^\infty \frac{\alpha b}{m\alpha+ab} \,\mathrm{B}\left(\frac{m\alpha}{a}, b\right)\label{eq:moment}
\end{aligned}
\eeq
where $\mathrm{B}(.,.)$ is the beta function and $b\,\mathrm{B}\left(\frac{m\alpha}{a}+1, b\right)$ is the $(m\, \alpha)^{\mathrm{th}}$ moment of the Kumaraswamy distribution with parameters $a$ and $b$. 
As the low-order moments dominate the infinite sum, we only use the first 10 terms to approximate (\ref{eq:moment}) in our experiments. Using (\ref{eq:kl1}) and (\ref{eq:kl2}) we have:
\begin{align}
\kl\left( q_v\left( a, b \right) || p_v\left( \alpha, \beta \right) \right) =& \log{ \frac{ab}{\alpha \beta} } - \frac{b-1}{b} + \frac{a-\alpha}{a} \left( -\gamma -\Psi(b) - \frac{1}{b} \right) + \sum_{m=1}^\infty \frac{\alpha b \left( \beta-1 \right) }{m \alpha + ab}\, \mathrm{B}\left(\frac{m\alpha}{a}, b\right) \label{eq:kl_kumar}
\end{align}
\section{Inference}
\label{apx:inf}
Recall that to determine which factors should be active for a particular task $t$, we perform variational inference to infer the posterior of parameters $\rvtheta_t = \{\rvb^\ell_t, \rvv^\ell_t \}_{\ell = 1}^L$.
The following variational distributions were used:
\begin{align}
    q(\rvtheta^\ell_t) &= q(\rvb_t^\ell) q(\rvv_t^\ell) \label{eq:post}\\
    \rvb_{t}^\ell &\sim \mathrm{Bernoulli}(\rvpi^\ell_t) \label{eq:app_post_b}\\
    \rvv_{t}^\ell &\sim \kumar(\rvc^\ell_t, \rvd^\ell_t)
\end{align}

The objective for each task is to maximize the variational bound:
\beq
\begin{aligned}
    \hspace{-0.5em} \mathcal L_t = \sum_{n=1}^{N_t} & \E_{q} \log{ p\left(y_t^{(n)} \big|\rvtheta_t, x_t^{(n)}, \rmW_a, \rmW_b, \rvr_t\right)}
    - \kl\left( q\left(\rvtheta_t\right)||p\left( \rvtheta_t \right)\right) \hspace{-1em} \label{eq:app_ELBO}
\end{aligned}
\eeq
where $N_t$ is the number of training examples in task $t$.
We use the mean-field approximation, so the second term can be expressed as

\begin{align}
    \kl\left( q(\rvtheta_t)||p\left( \rvtheta_t \right)\right)
    =&  \E_{q(\rvv^\ell_t)}\left[\kl\left( q(\rvb^\ell_t) || p(\rvb^\ell_t|\rvv^\ell_t) \right)\right]
    + \kl\left( q(\rvv^\ell_t) || p(\rvv^\ell_t) \right) \label{eq:app_online_inference}
\end{align}
% \beq
% \begin{aligned}
%     \kl\left( q(\rvtheta_t)||p\left( \rvtheta_t \right)\right)
%     =&  \E_{q(\rvv^\ell_t)}\left[\kl\left( q(\rvb^\ell_t) || p(\rvb^\ell_t|\rvv^\ell_t) \right)\right]
%     + \kl\left( q(\rvv^\ell_t) || p(\rvv^\ell_t) \right) \label{eq:app_online_inference}
% \end{small}
% \end{aligned}
% \eeq
where the first term in (\ref{eq:app_online_inference}) is approximated by taking samples from $\rvv_{t}^\ell \sim q(\rvv_{t}^\ell)$.

VCL \citep{Nguyen2018} addresses catastrophic forgetting using online inference, $i.e.$, the posterior inferred from the most recent task is used as a prior for the incoming task. However, more recent work~\citep{farquhar2018robust, farquhar2019unifying} suggests that online inference often does not succeed in mitigating catastrophic forgetting in realistic continual learning settings, as methods based solely on online inference rely on the prior capturing everything learned on all previous tasks. Thus in (\ref{eq:app_online_inference}), instead of performing online inference for $\{\rvb_t^\ell, \rvv^\ell_t\}$, we only apply online inference for $\vv^\ell_t$, while learning task-specific parameters $\{\rvr^\ell_t, \rvb_t^\ell\}$. 

\paragraph{Inference for $\rvv^\ell_t$:} Starting with the first task ($t=1$), we initialize the prior $p(\rvv^\ell_1) = \mathrm{Beta}(\alpha, 1)$ and learn the posterior $q(\rvv^\ell_1) = \kumar(\rvc^\ell_1, \rvd^\ell_1)$ using Bayes by Backprop \citep{blundell2015}. Note that $\mathrm{Beta}(\alpha, 1)$ has the same density function as $\kumar (\alpha, 1)$. For all the following tasks, the prior $p(\rvv^\ell_t) = q(\rvv^\ell_{t-1})$ and the posterior $q(\rvv^\ell_{t}) = \kumar(\rvc^\ell_t, \rvd^\ell_t)$ is learned in the same way as in task 1. Note that we use mean-field approximation for the posterior: $q(\ervv^\ell_{t,i}) = \kumar(\evc^\ell_{t,i}, \evd^\ell_{t,i})$. We use (\ref{eq:kl_kumar}) to compute the KL divergence between the posterior and the prior in (\ref{eq:app_online_inference}).

\paragraph{Inference for $\rvb_t^\ell$:} We use the ${\mathrm{BernoulliConcrete}_\lambda}$ distribution \citep{MaddisonMT16} as the soft approximation of the Bernoulli distribution for both the prior and the posterior. We fix $\lambda = 2/3$ for all our experiments. We employ the prior $p(\ervb_{t,k}^\ell) = \mathrm{BernoulliConcrete}_\lambda({\ervpi^{\ell}_{t,k}})$, where ${\ervpi^{\ell}_{t,k}} := \prod_{i=1}^{i=k}\ervv^{\ell}_{t,i}$ and $\ervv^{\ell}_{t,i} \sim q(\ervv^\ell_{t,i})$. The posterior is then $q(\ervb^\ell_{t,k}) = \mathrm{BernoulliConcrete}_\lambda({\bar{\pi}^{\ell}_{t,k}})$, where ${\bar{\pi}^{\ell}_{t,k}}$ is learned using Bayes by Backprop. We use the the \ac{KL} divergence and reparameterization for the $\mathrm{BernoulliConcrete}_\lambda$ as given by \cite{MaddisonMT16}.

\section{Task Inference at Test Time}
Recall from (\ref{eq:phi_task_inference}), we approximate $P(t|x)$ with $P(t|\phi(x))$,
where:
\begin{align}
  P(t|x) \approx \frac{P(\phi(x)|t) P(t)}{\sum_{t'} P(\phi(x)|t') P(t')} = P(t|\phi(x))
  \label{apx_eq:phi_task_inference}
\end{align}
In the following section, we show that under a certain assumption (namely Assumption \ref{apx_assumption:1} in \ref{apx:task_infer_proof}), this approximation is exact with $P(t|x) = P(t|\phi(x))$. However, in practice this assumption may not hold without an explicit hard constraint; hence we consider (\ref{eq:phi_task_inference}) an approximation. 
Nevertheless, we feel it is important to show this connection.
In section~\ref{apx:task_results}, we show empirical results on employing task inference as described in (\ref{apx_eq:phi_task_inference}) over commonly used continual learning benchmarks.
 
\subsection{Proof}
\label{apx:task_infer_proof}
Let $\phi: \mathcal{X} \rightarrow \mathcal{S}$ be the transformation function. We will assume $\phi$ is differentiable. For the transformation $\phi: \mathcal{X} \rightarrow \mathcal{S}$, and for a distribution $P$ defined over $\mathcal{X}$, let $P_\phi$ be the distribution induced by $\phi$ over $\mathcal{S}$.

\begin{assumption}
\label{apx_assumption:1}
The transformation $\phi$ is a one-to-one function. Without loss of generality assume $\mathcal{S}$ to be the image of $\mathcal{X}$ under $\phi$ with $\psi: \mathcal{S} \rightarrow \mathcal{X}$ to be the inverse of $\phi$, such that $\psi\left(\phi(x)\right)= x$.
 \end{assumption}

{\begin{lemma} (Remark, main text).
    \label{apx_lemma_1}
    If Assumption~\ref{apx_assumption:1} holds, then $P(t|\phi(x)) = P(t|x)$ $\forall \, x \in \mathcal{X}$, $t \in \{1,2,...T\}$.
\end{lemma}
$Proof.$ Let $M_\phi(s) = \left|\mathrm{det} \frac{\partial \phi^{-1}(s)}{\partial s}\right|$ be the absolute of the determinant of the Jacobian of $\psi(s)$. Consider $s = \phi(x)$ and $x = \psi(s)$.\\
    \begin{align}
        P_\phi(t|s) &= \frac{P_\phi(t,s)}{P_\phi(s)}\\ &\overset{(a)}{=} \frac{P(t,\psi(s))\, M_\phi(s)}{P(\psi(s))\, M_\phi(s)}\\
        &= \frac{P(t,x)}{P(x)} = P(t|x)
    \end{align}
    where (a) follows from the change of variable formula. Note that $M_\phi(s)$ can also be written as $\left|\mathrm{det} \frac{\partial \phi(x)}{\partial x}\right|^{-1}$ if $\psi$ is continuously differentiable (it is not however a requirement for Lemma $\ref{apx_lemma_1}$).
}

\subsection{Quantitative Results for Task
Inference and Visualizations}
\label{apx:task_results}
\begin{wrapfigure}{r}{0.5\textwidth}
    \vspace{-0.5em}
    \centering
    \includegraphics[width=0.4\textwidth]{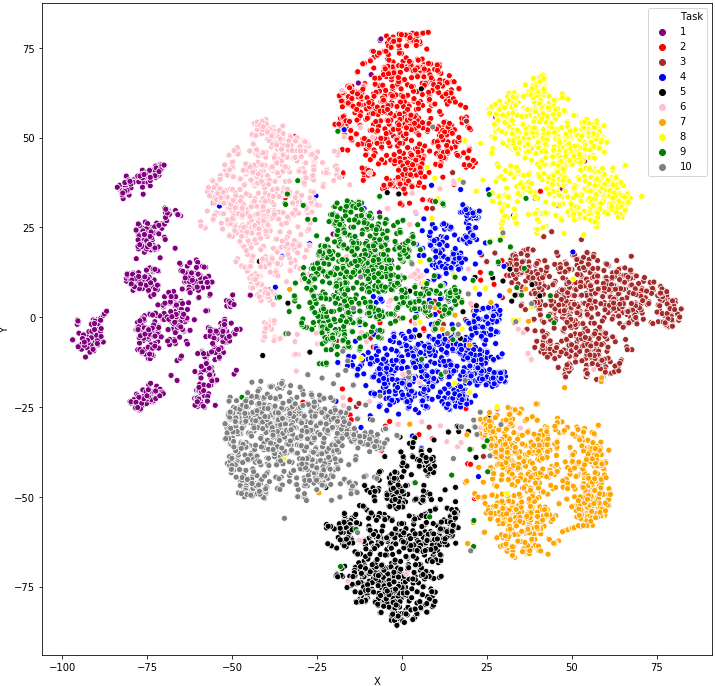}
    \caption{The t-SNE plot for the intermediate features $\phi$ of 10 tasks from the permuted MNIST benchmark.}
    \label{fig:permute_tsne}
\end{wrapfigure}
The procedure introduced in Section~\ref{sec:task_infer} is used for identifying the task identity during evaluation when it is otherwise unavailable.
Thus, incremental class performance is highly dependent on task inference accuracy.
We report the task inference accuracy in Table~\ref{apx_tbl:task_inference}. 

Additionally, we visualize with a t-SNE plot the distribution of the intermediate features $\phi$ from 10 tasks of permuted MNIST in Figure~\ref{fig:permute_tsne}.
The features across tasks are noticeably clustered, which allows our task inference method to infer task identity from simple feature statistics. While the accuracy for CIFAR10 is much lower than for MNIST, this is partially attributable to the inherent challenge of sequentially learned task inference for CIFAR10: CIFAR10 proves challenging for the generative models commonly used by replay methods for task inference as well. 
For example, we find that learning a separate VAE~\citep{kingma2013autoencoding}\footnote{For VAE task inference, we used an encoder with layers 3(input)-32(conv)-64(conv)-128(fc), with a decoder that had a deconvolution architecture symmetrical to the encoder. We used the ELBO to approximate $P(x|t)$.} for each task resulted in a task inference accuracy of $~41\%$. In general, \cite{nalisnick2018deep} showed that the density estimates from generative models can lead to poor Out-of-Distribution detection. However, a comprehensive study is required for further analysis. We leave further exploration of generative models for task inference for future work.
\begin{table}[h]
\centering
\caption{\label{apx_tbl:task_inference}
The task inference accuracy using (\ref{apx_eq:phi_task_inference}).}
\begin{tabular}{ccc}
\toprule
Split MNIST & Permuted MNIST & CIFAR10\\ 
\midrule
92.63 $\pm$ 0.12 & 99.98 $\pm$ 0.01 & 43.62 $\pm$ 0.16\\
\bottomrule
\end{tabular}% 0.1 & 0.2 & 0.3 & 0.4 & 0.5 & 0.6 & 0.7 & 0.8 & 0.9 & 0.9 & 1.0
\end{table}

\section{Uncertainty Estimation}
\label{apx:uncertainty}
A desired behaviour from a model is to return the uncertainty (or confidence) associated each prediction. Neural networks are prone to have high confidence when the input lies outside of the training distribution. For such inputs, we want our model to have high uncertainty (or low confidence) associated with the predictions. Unlike neural networks trained as point-estimates (using MLE/MAP), Bayesian neural networks provide a natural framework to estimate uncertainty associated with the prediction. We estimate the uncertainty in a continual setting for both incremental task and incremental class settings. Note that non-Bayesian continual learning methods do not have principled method to estimate uncertainty. Our estimate of uncertainty is based on the predictive entropy defined as:
\begin{align}
\mathbb{H}\left[y^*|x^*, \cD_{train}\right] = -\sum_{c} \Big( &p(y^*=c|x^*, \cD_{train}) \log{p(y^*=c|x^*, \cD_{train})} \Big)
\end{align}
\begin{figure*}[!h]
    \centering
    \includegraphics[width=\textwidth]{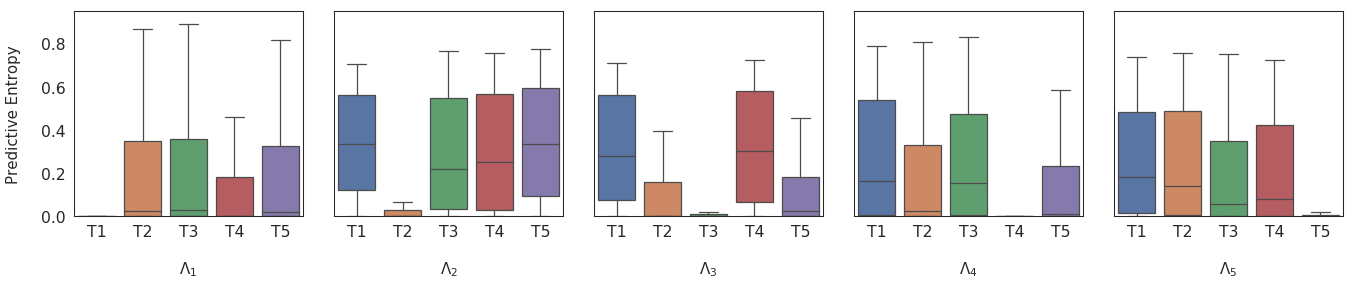}
    \vspace{-1em}
    \caption{(Section~\ref{apx:task_uncertainty}) Uncertainty in the incremental task setting for Split MNIST dataset. Each of the 5 plots depicts the uncertainxty of the test sets when task-specific parameter $\Lambda_t$ is used. The $y$-axis denotes the uncertainty (as the predictive entropy), and $x$-axis denotes the test sets ($\mathcal{T}_1$ through $\mathcal{T}_5$).}
    \label{fig:incremental_task}
\end{figure*}
\begin{figure*}[ht]
    \centering
    \includegraphics[width=\textwidth]{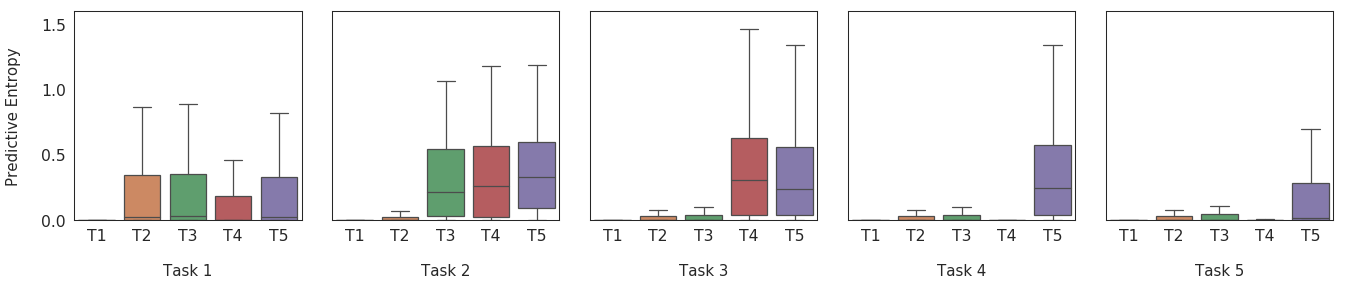}
    \caption{(Section~\ref{apx:class_uncertainty}) Uncertainty in the incremental class setting for Split MNIST dataset. We compute the uncertainty of the test sets after training on each task in the sequence. The $y$-axis denotes the uncertainty (as the predictive entropy), and $x$-axis denotes the test sets ($\mathcal{T}_1$ through $\mathcal{T}_5$) for each task. Since we do not know the task (and the corresponding $\Lambda_t$), the predictive entropy is computed by marginalizing over all tasks. All the seen classes have low uncertainty compared to unseen ones.}
    \label{fig:incremental_class}
\end{figure*}
\subsection{Incremental Task Learning:} 
\label{apx:task_uncertainty}
Recall that in incremental task learning, we know the task identity at test time. Hence, we compute the predictive distribution by doing a forward pass using the task-specific parameters in IBP-WF; we can write $p(y^*=c|x^*) = p(y^*=c|x^*, t^*)$, where $t^*$ is the associated task-identity with the input $x^*$ during testing. Following \cite{Gal2016Uncertainty}, we approximate the predictive distribution by using an ensemble of $M$ neural networks sampled from the posterior distribution:
\begin{align}
    p(y^*=c|t=t^*,x^*) = \underbrace{\frac{1}{M}\sum_{m=1}^M p(y^*=c|x^*; \theta^{(m)}_{t^*})}_{\rho^{M}_{t^*,c}}, \enspace \text{where} \enspace \theta^{(m)}_{t^*} \sim q(\theta_{t^*})
\end{align}
where $\theta^{(1)}...\theta^{(M)}$ are $M$ samples drawn from $q(\theta_{t^*})$. Using this we can compute a biased estimate\footnote{The estimate is biased since $\mathbb{H[.]}$ is a non-linear function. The bias will decrease as $M$ increases.} of the predictive entropy as follows:
\begin{align}
    \mathbb{H}\left[y^*|x^*\right] &= -\sum_{c} p(y^*=c|x^*) \log{p(y^*=c|x^*)} \\
    &= -\sum_{c} p(y^*=c|x^*, t^*) \log{p(y^*=c|x^*, t^*)} \\
    &= -\sum_{c} \left( \frac{1}{M}\sum_{m=1}^M p(y^*=c|x^*; \theta^{(m)}_{t^*}) \right) \log{\left( \frac{1}{M}\sum_{m=1}^M p(y^*=c|x^*; \theta^{(m)}_{t^*}) \right)}\\
    &= -\sum_{c}  \left( \rho^{M}_{t^*,c} \right) \log{\rho^{M}_{t^*,c}}
\end{align}
Figure~\ref{fig:incremental_task} shows the uncertainty estimates for the test sets in the Split MNIST dataset. We denote the test set for a task $t \in \{1...5\}$ as $\mathcal{T}_t$. As it can be seen in Figure~\ref{fig:incremental_task}, given a task-identity $t$, the uncertainty for the test set $\mathcal{T}_t$ when used with parameters $\Lambda_t$ is significantly smaller compared to the uncertainty of test sets $\{\mathcal{T}_t' \, | \, t' \neq t\}$. One application of computing uncertainties would be an out-of-distribution test in the continual learning setting. However, we leave exploring such extensions for future work. We use $M=100$ to compute the uncertainty.

\subsection{Incremental Class Learning:} 
\label{apx:class_uncertainty}
For the incremental class setting, we do not have access to the task-identity of a given test point. We use the task inference mechanism from Section~2.4 in the main paper. To infer the predictive distribution, we marginalize over the task-identities:
\begin{align}
    p(y^*=c|x^*) &= \sum_{t'} p(y^*=c, t=t'|x^*) \\
    &= \sum_{t'} p(y^*=c|t=t',x^*) p(t=t'| x^*) \\
    &\approx \sum_{t'} p(y^*=c|t=t',x^*)\frac{P(\phi(x)|t') P(t')}{\sum_t P(\phi(x)|t)P(t) }\\
    &= \sum_{t'} \rho^{M}_{t',c} \, P(t'|\phi(x)) 
\end{align}
\begin{align}
    \label{eq:entropy_incr_class}
    \mathbb{H}\left[y^*|x^*\right] &= -\sum_{c} \left( \left( \sum_{t'} \rho^{M}_{t',c} \, P(t'|\phi(x)) \right) \log{ \left( \sum_{t'} \rho^{M}_{t',c} \, P(t'|\phi(x)) \right) } \right) 
\end{align}

We use (\ref{eq:entropy_incr_class}) to estimate the uncertainty in the incremental class continual setting for the Split MNIST dataset. Figure~\ref{fig:incremental_class} shows the uncertainty of test sets after training on each task. As shown, initially when the model is trained on the first task, the uncertainty of $\mathcal{T}_2$-$\mathcal{T}_5$ is higher than the uncertainty of $\mathcal{T}_1$. As the training progresses, the uncertainty of the corresponding task decreases while still maintaining a low estimate of the uncertainty of the test sets for the previous tasks. This provides further evidence that our proposed method IBP-WF mitigates catastrophic forgetting.  We use $M=100$ to compute the uncertainty.

\section{Baselines}
\label{apx:baselines}
We compare IBP-WF with a number of other approaches outlined as follows:

\textbf{Fine-tuning} The model is trained by a stochastic gradient descent algorithm, seeing each task in sequence. 
At the conclusion of each task, the ``final'' trained model for a task is used as the initialization for the next task.
This represents the naive approach to training on sequential task data, where catastrophic forgetting was first recognized.
We compare against models trained by vanilla stochastic gradient descent (\textbf{SGD}) with constant learning rate, as well as by adaptive learning rate methods \textbf{Adam}~\citep{Kingma2014} and \textbf{Adagrad}~\citep{Duchi2011}.

\textbf{Regularization Methods} Recognizing that training on a new task may result in a model's parameters moving away from an optimum for a previous task, a number of continual learning strategies attempt to constrain the model parameters from drifting too far while learning a new task.
A simple way to do so is to apply an $L_2$ loss on the model parameters' distance from previous task solutions.
\textbf{EWC}~\citep{Kirkpatrick2017} refines this by weighting the $L_2$ by parameter importance, using the Fisher Information; \textbf{Online EWC} \citep{Schwarz2018, Liang2018b} uses an online version that provides better scaling.
\textbf{SI}~\citep{Zenke2017} also weights the $L_2$ regularization by importance, with the importance weighting instead coming from the amount a parameter contributed to reducing the loss over its trajectory.
\textbf{MAS}~\citep{Aljundi2018} computes parameter importance as well, but with respect to the model output rather than the loss.
\textbf{LwF}~\citep{Li2017} leverages knowledge distillation~\citep{Hinton2015} principles, using previous model outputs as additional training objectives.
\textbf{VCL} uses Bayesian neural networks, using the posterior of the previous task as the prior for the next. We also compare against a recent expansion method called \textbf{IBNN} \citep{kessler2020hierarchical} that uses \ac{IBP} to adapt the structure of a Bayesian neural network. The accuracy for IBNN is taken directly from \cite{kessler2020hierarchical}.\footnote{The IBNN performance on permMNIST is based on only 5 tasks, whereas other methods in Table~\ref{tbl:additional_mnist_exp} use 10 tasks.}

\textbf{Replay Methods} 
As catastrophic forgetting can be attributed to not seeing previous parts of the data distribution, another class of methods employ experience replay: refreshing the model on old tasks while learning new ones.
\textbf{Naive Rehearsal} accomplishes this by keeping examples from old tasks in a buffer and assembling them into ``replay'' minibatches.
This runs the risk of overfitting the samples in the buffer, so \textbf{GEM}~\citep{Lopez2017} proposes instead using these as inequality restraints: the model should not increase the loss on saved samples.
These saved samples can also used for re-training or fine-tuning the model, which \textbf{VCL}~\citep{Nguyen2018} does with its coresets.
Regardless of how stored samples are used, however, in certain settings, data is private~\citep{Ribli2018} or classified~\citep{Liang2018a}, and keeping data may be considered as violating continual learning criteria.
As an alternative, \textbf{DGR}~\citep{Shin2017} and \textbf{RtF}~\citep{Van2018} propose generative models \citep{Goodfellow2014} as a source of replay.
Such approaches avoid carrying around older data, but require learning (and storing) generative models for each task, which may need to be quite large depending on the complexity of the dataset.

We use the codebase from \citep{Hsu2018, vandeven2019three} as our continual learning ``sandbox.'' Best efforts were made to keep the model capacity consistent in all methods for a fair comparison. 

\begin{table*}[!h]
\centering
% \scriptsize
\caption{\label{tbl:additional_mnist_exp}
\small The average accuracy of seen tasks after learning on a sequence of tasks using a \ac{MLP}.}
\resizebox{0.8\textwidth}{!}{%
\begin{tabular}{lcccc}
\toprule
Method & \multicolumn{2}{c}{Split MNIST} & \multicolumn{2}{c}{Permuted MNIST} \\ \midrule
& Incremental Task & Incremental Class & Incremental Task & Incremental Class \\ \toprule
SGD & 96.95 $\pm$ 0.46 & 19.46 $\pm$ 0.04 & 90.54 $\pm$ 0.03 & 8.46 $\pm$ 0.36 \\
Adam & 95.18 $\pm$ 2.64 & 19.71 $\pm$ 0.08 & 91.70 $\pm$ 1.89 & 16.13 $\pm$ 0.71\\
$L_2$ & 98.32 $\pm$ 0.73 & 22.52 $\pm$ 1.08 & 94.01 $\pm$ 0.27 & 16.43 $\pm$ 0.63 \\ 
Online EWC & 99.09 $\pm$ 0.12 & 19.77 $\pm$ 0.04 & 93.62 $\pm$ 0.25 & 42.40 $\pm$ 2.68 \\ 
IBNN & 95.30 $\pm$ 2.00 & 85.50 $\pm$ 3.20 & 95.6\footnotemark[3] $\pm$ 0.20 & 93.8\footnotemark[3] $\pm$ 0.30 \\
\midrule
\textbf{IBP-WF} (Ours) & \textbf{99.69} $\pm$ 0.05 & 92.40 $\pm$ 0.15 & \textbf{97.52} $\pm$ 0.06 & \textbf{97.50} $\pm$ 0.06 \\ 
\bottomrule
\end{tabular}
}
\end{table*}

\section{Experiment Setup}
\label{apx:setup}
We describe the experimental configuration used:

\subsection{Split MNIST}
Following \citep{Hsu2018}, we use the standard train/test split, with 60K training images (6K images per digit) and 10K test images (1K images per digit). 
Standard normalization of the images was the only preprocessing done, without any data augmentation strategies used for any of the algorithms.

\paragraph{Baselines:} All baseline methods use the same neural network architecture: a \ac{MLP} with two hidden layers of 400 nodes each, followed by a softmax layer. For GEM and naive rehearsal, a buffer of 400 images were saved to replay previous tasks. For DGR and RtF a 2-layer symmetric variational autoencoder\citep{kingma2013autoencoding} was learned for each task. We used ReLU as the non-linearity in both the hidden layers. All the baseline models, except VCL, RtF and HIBNN, were trained for 10 epochs per task with a mini-batch size of 128 with Adam~\citep{Kingma2014} optimizer ($\beta_1 = 0.9, \beta_2=0.999,  lr=0.001$) as the default unless explicitly stated. VCL was trained for 50 epochs. The results for RtF were taken from the original paper by \cite{Van2018}, which was trained for 2000 steps with a batch size of 128. Note that RtF has twice the number of parameters compared to IBP-WF. The results for HIBNN in table~\ref{tbl:additional_mnist_exp} were taken from \cite{kessler2020hierarchical}, which was trained for 200 epochs. For EWC online, EWC, SI, GEM and MAS, the regularization coefficient was set to 400, 100, 300, 0.5 and 1.0 respectively.

\paragraph{IBP-WF:} IBP-WF used the same neural architecture as the baselines, except there was only a single hidden layer. The prior parameter for \ac{IBP} was set to $\alpha=100$. The model is expanded for 10 epochs (using equation~(\ref{eq:ELBO}) in the main paper) with a learning rate of 0.001 and fine-tuned with a fixed number of factors for 5 epochs. A mini-batch size of 32 was used. We used the stick-breaking construction for IBP, which was truncated at $K=400$, $i.e.$ the total budget on the number of allowed factors was 400.

\subsection{Permuted MNIST}
We use the standard train/test split of the MNIST dataset. Each task consists of the same 10-way digit classification, but with the pixels of the entire MNIST dataset randomly permuted in a consistent manner. We generate 10 such tasks using 10 random permutations in our experiments.

\paragraph{Baselines:} All the baseline methods use the same neural network architecture: a \ac{MLP} with two hidden layers of 1000 nodes each, followed by a softmax layer. We used ReLU as the non-linearity in both the hidden layers. For GEM and naive rehearsal, a buffer 1.1K images were saved to replay previous tasks. For DGR and RtF a 2-layer symmetric variational autoencoder was learned for each task. All the baseline models, except RtF and VCL, were trained for 15 epochs per task with a mini-batch size of 128 with Adam optimizer ($\beta_1 = 0.9, \beta_2=0.999,  lr=0.001$) as the default unless explicitly stated. For VCL, the model was trained for 100 epochs. The results for RtF were taken from \cite{Van2018}, which was trained for 5000 iterations. For EWC online, EWC, SI, GEM and MAS, the regularization coefficient was set to 500, 500, 1.0, 0.5 and 0.01 respectively.

\paragraph{IBP-WF}
IBP-WF used the same neural architecture as the baselines. The prior parameter was set to $\alpha=700$. We train for 15 epochs for each task using \ac{IBP} (using equation (\ref{eq:ELBO}) in the main paper) with a mini-batch size of 64. The model was then fine-tuned for 5 epochs with a fixed number of factors. The stick-breaking process for IBP was truncated at $K=1000$ for both the hidden layers.

\subsection{CIFAR10}
We split the CIFAR10~\citep{Krizhevsky2009} dataset into a sequence of 5 binary classification tasks.  Similar to Split MNIST, this is a binary classification problem at test time in the incremental task setting, and 10-way classification in the incremental class setting.

\paragraph{Baselines:} We use ResNet-20~\citep{He2016} for all the baselines. We used standard data augmentation methods (random crop, horizontal flips and standard normalization) while training. All the baselines models were trained for 160 epochs per task with a mini-batch size of 128. A learning rate of $lr = 0.001$ was used. For naive rehearsal, a buffer of 400 images were saved to replay previous tasks. For EWC online, EWC, and SI, the regularization coefficient was set to 3000, 100 and 2 respectively.
\paragraph{IBP-WF:} We scale our IBP-WF method to ResNet-20 by factorizing convolutional layers. The training was carried out for 160 epochs with a learning rate of $lr=0.001$ and a mini-batch size of 128. There was no fine-tuning done for this experiment. The truncation parameters for the stick-breaking process was set to $200$ for all the layers. We used task-specific batch normalization parameters for our implementation. We set the IBP hyperparameter $\alpha$ to be $40$ for all the convolutional layers and $32$ for the final fully-connected layer.

\vspace{-2mm}
\section{Ablation Studies}

\subsection[Selecting alpha]{Selecting $\alpha$}
\label{apx:ablation_alpha}
\begin{figure}[!h]
    \centering
    \includegraphics[width=0.8\textwidth]{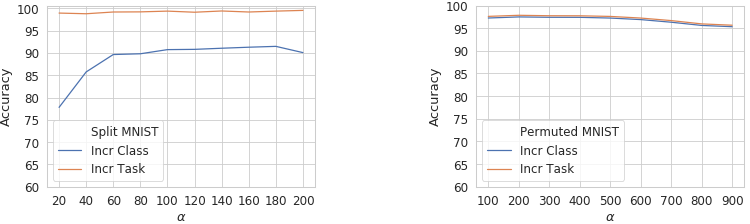}
    \caption{Ablation study on $\alpha$.}
    \label{fig:ablation_alpha}
\end{figure}

The hyperparameter $\alpha$ controls the behavior of the Indian Buffet Process prior, which for IBP-WF provides a regularization effect for both the number of active (nonzero) factors per task, as well as the expected rate at which new factors are added (expansion).
Specifically, $\alpha$ \textit{is} the prior's expected number of factors per task, and as such should be a value on the order of (but preferably less than) the rank of the weight matrix.
We sweep $\alpha$ and plot overall final performance of IBP-WF on Split MNIST and Permuted MNIST in both incremental class and incremental task settings in Figure~\ref{fig:ablation_alpha}.
We observe that excessively low values of $\alpha$ lead to poorer performance, as there are not enough factors to learn each task, but otherwise IBP-WF exhibits low sensitivity to $\alpha$ over a very wide range of values, showing relative robustness to $\alpha$.

\subsection[Selecting kappa]{Selecting $\kappa$}
\label{apx:ablation_kappa}
\begin{figure}[!h]
    \centering
    \includegraphics[width=0.8\textwidth]{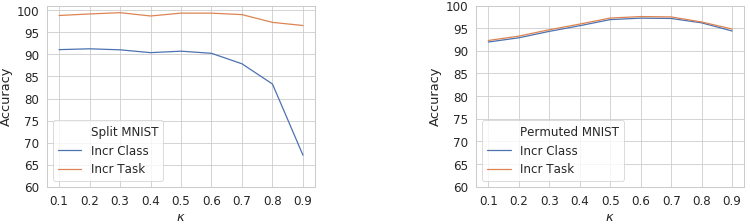}
    \caption{Ablation study on $\kappa$. We use $\kappa=0.5$ for all experiments in the main text.}
    \label{fig:ablation_kappa}
\end{figure}

IBP-WF preserves past knowledge by selectively freezing weight factors that played a key role in previous tasks.
We define this criterion as factors whose probability $\pi_{t,k}^\ell$ exceed a threshold $\kappa$.
As with $\alpha$, we sweep $\kappa$ and plot IBP-WF's performance on Split MNIST and Permuted MNIST in both incremental class and incremental task settings in Figure~\ref{fig:ablation_kappa}.
We observe a decline in performance if $\kappa$ is set too high for incremental class learning in Split MNIST, as it likely leads to not enough factors being preserved, but overall there is a wide range of settings of $\kappa$ that give good performance.

{
\vfill
\small
\begin{@fileswfalse}
\bibliography{bibliography}

\begin{thebibliography}{55}
\providecommand{\natexlab}[1]{#1}
\providecommand{\url}[1]{\texttt{#1}}
\expandafter\ifx\csname urlstyle\endcsname\relax
  \providecommand{\doi}[1]{doi: #1}\else
  \providecommand{\doi}{doi: \begingroup \urlstyle{rm}\Url}\fi

\bibitem[Aljundi et~al.(2018)Aljundi, Babiloni, Elhoseiny, Rohrbach, and
  Tuytelaars]{Aljundi2018}
Rahaf Aljundi, Francesca Babiloni, Mohamed Elhoseiny, Marcus Rohrbach, and
  Tinne Tuytelaars.
\newblock {Memory Aware Synapses: Learning What (not) to Forget}.
\newblock \emph{European Conference on Computer Vision}, 2018.

\bibitem[Aljundi et~al.(2019)Aljundi, Kelchtermans, and
  Tuytelaars]{Aljundi2019b}
Rahaf Aljundi, Klaas Kelchtermans, and Tinne Tuytelaars.
\newblock {Task-Free Continual Learning}.
\newblock \emph{Computer Vision and Pattern Recognition}, 2019.

\bibitem[Blundell et~al.(2015)Blundell, Cornebise, Kavukcuoglu, and
  Wierstra]{blundell2015}
Charles Blundell, Julien Cornebise, Koray Kavukcuoglu, and Daan Wierstra.
\newblock {Weight Uncertainty in Neural Networks}.
\newblock \emph{International Conference on Machine Learning}, 2015.

\bibitem[Duchi et~al.(2011)Duchi, Hazan, and Singer]{Duchi2011}
John Duchi, Elad Hazan, and Yoram Singer.
\newblock {Adaptive Subgradient Methods for Online Learning and Stochastic
  Optimization}.
\newblock \emph{Journal of Machine Learning Research}, 12\penalty0
  (Jul):\penalty0 2121--2159, 2011.

\bibitem[Farquhar \& Gal(2018)Farquhar and Gal]{farquhar2018robust}
Sebastian Farquhar and Yarin Gal.
\newblock {Towards Robust Evaluations of Continual Learning}.
\newblock \emph{arXiv preprint arXiv:1805.09733}, 2018.

\bibitem[Farquhar \& Gal(2019)Farquhar and Gal]{farquhar2019unifying}
Sebastian Farquhar and Yarin Gal.
\newblock {A Unifying Bayesian View of Continual Learning}.
\newblock \emph{arXiv preprint arXiv:1902.06494}, 2019.

\bibitem[Gal(2016)]{Gal2016Uncertainty}
Yarin Gal.
\newblock \emph{Uncertainty in Deep Learning}.
\newblock PhD thesis, University of Cambridge, 2016.

\bibitem[Ghahramani \& Griffiths(2006)Ghahramani and Griffiths]{Ghahramani2006}
Zoubin Ghahramani and Thomas~L Griffiths.
\newblock {Infinite Latent Feature Models and the Indian Buffet Process}.
\newblock \emph{Neural Information Processing Systems}, 2006.

\bibitem[Goodfellow et~al.(2014)Goodfellow, Pouget-Abadie, Mirza, Xu,
  Warde-Farley, Ozair, Courville, and Bengio]{Goodfellow2014}
Ian Goodfellow, Jean Pouget-Abadie, Mehdi Mirza, Bing Xu, David Warde-Farley,
  Sherjil Ozair, Aaron Courville, and Yoshua Bengio.
\newblock {Generative Adversarial Nets}.
\newblock \emph{Neural Information Processing Systems}, 2014.

\bibitem[Goodfellow et~al.(2013)Goodfellow, Mirza, Xiao, Courville, and
  Bengio]{Goodfellow2013}
Ian~J Goodfellow, Mehdi Mirza, Da~Xiao, Aaron Courville, and Yoshua Bengio.
\newblock {An Empirical Investigation of Catastrophic Forgetting in
  Gradient-based Neural Networks}.
\newblock \emph{arXiv preprint arXiv:1312.6211}, 2013.

\bibitem[Gradshteyn \& Ryzhik(2007)Gradshteyn and Ryzhik]{gradshteyn2007}
I.~S. Gradshteyn and I.~M. Ryzhik.
\newblock \emph{Table of integrals, series, and products}.
\newblock Elsevier/Academic Press, Amsterdam, seventh edition, 2007.

\bibitem[He et~al.(2016)He, Zhang, Ren, and Sun]{He2016}
Kaiming He, Xiangyu Zhang, Shaoqing Ren, and Jian Sun.
\newblock {Deep Residual Learning for Image Recognition}.
\newblock \emph{Computer Vision and Pattern Recognition}, 2016.

\bibitem[Heusel et~al.(2017)Heusel, Ramsauer, Unterthiner, Nessler, and
  Hochreiter]{heusel2017gans}
Martin Heusel, Hubert Ramsauer, Thomas Unterthiner, Bernhard Nessler, and Sepp
  Hochreiter.
\newblock {GANs Trained by a Two Time-Scale Update Rule Converge to a Local
  Nash Equilibrium}.
\newblock \emph{Neural Information Processing Systems}, 2017.

\bibitem[Hinton et~al.(2015)Hinton, Vinyals, and Dean]{Hinton2015}
Geoffrey Hinton, Oriol Vinyals, and Jeff Dean.
\newblock {Distilling the Knowledge in a Neural Network}.
\newblock \emph{arXiv preprint arXiv:1503.02531}, 2015.

\bibitem[Hochreiter \& Schmidhuber(1997)Hochreiter and
  Schmidhuber]{Hochreiter1997lstm}
Sepp Hochreiter and J{\"{u}}rgen Schmidhuber.
\newblock {Long Short-Term Memory}.
\newblock \emph{Neural Computation}, 1997.

\bibitem[Hsu et~al.(2018)Hsu, Liu, Ramasamy, and Kira]{Hsu2018}
Yen-Chang Hsu, Yen-Cheng Liu, Anita Ramasamy, and Zsolt Kira.
\newblock {Re-evaluating Continual Learning Scenarios: A Categorization and
  Case for Strong Baselines}.
\newblock \emph{arXiv preprint arXiv:1810.12488}, 2018.

\bibitem[Hung et~al.(2019)Hung, Tu, Wu, Chen, Chan, and Chen]{Hung2019}
Ching-Yi Hung, Cheng-Hao Tu, Cheng-En Wu, Chien-Hung Chen, Yi-Ming Chan, and
  Chu-Song Chen.
\newblock {Compacting, Picking and Growing for Unforgetting Continual
  Learning}.
\newblock \emph{Neural Information Processing Systems}, 2019.

\bibitem[Kessler et~al.(2020)Kessler, Nguyen, Zohren, and
  Roberts]{kessler2020hierarchical}
Samuel Kessler, Vu~Nguyen, Stefan Zohren, and Stephen Roberts.
\newblock Hierarchical indian buffet neural networks for bayesian continual
  learning.
\newblock \emph{arXiv preprint arXiv:1912.02290}, 2020.

\bibitem[Kingma \& Ba(2014)Kingma and Ba]{Kingma2014}
Diederik~P Kingma and Jimmy Ba.
\newblock {Adam: A Method for Stochastic Optimization}.
\newblock \emph{arXiv preprint arXiv:1412.6980}, 2014.

\bibitem[Kingma \& Welling(2013)Kingma and Welling]{kingma2013autoencoding}
Diederik~P Kingma and Max Welling.
\newblock Auto-encoding variational bayes, 2013.

\bibitem[Kirkpatrick et~al.(2017)Kirkpatrick, Pascanu, Rabinowitz, Veness,
  Desjardins, Rusu, Milan, Quan, Ramalho, Grabska-Barwinska, Hassabis, Clopath,
  Kumaran, and Hadsell]{Kirkpatrick2017}
James Kirkpatrick, Razvan Pascanu, Neil Rabinowitz, Joel Veness, Guillaume
  Desjardins, Andrei~A Rusu, Kieran Milan, John Quan, Tiago Ramalho, Agnieszka
  Grabska-Barwinska, Demis Hassabis, Claudia Clopath, Dharshan Kumaran, and
  Raia Hadsell.
\newblock {Overcoming Catastrophic Forgetting in Neural Networks}.
\newblock \emph{Proceedings of the National Academy of Sciences}, 2017.

\bibitem[Krizhevsky(2009)]{Krizhevsky2009}
Alex Krizhevsky.
\newblock {Learning Multiple Layers of Features from Tiny Images}.
\newblock 2009.

\bibitem[Kumar et~al.(2019)Kumar, Chatterjee, and Rai]{Kumar2019}
Abhishek Kumar, Sunabha Chatterjee, and Piyush Rai.
\newblock {Nonparametric Bayesian Structure Adaptation for Continual Learning}.
\newblock \emph{arXiv preprint arXiv:1912.03624}, 2019.

\bibitem[Kumaraswamy(1980)]{Kumaraswamy1980}
Ponnambalam Kumaraswamy.
\newblock {A Generalized Probability Density Function for Double-Bounded Random
  Processes}.
\newblock \emph{Journal of Hydrology}, 1980.

\bibitem[LeCun et~al.(1998)LeCun, Bottou, Bengio, and Haffner]{Lecun1998}
Yann LeCun, L{\'e}on Bottou, Yoshua Bengio, and Patrick Haffner.
\newblock {Gradient-based Learning Applied to Document Recognition}.
\newblock \emph{Proceedings of the IEEE}, 86\penalty0 (11):\penalty0
  2278--2324, 1998.

\bibitem[Lee et~al.(2018)Lee, Lee, Lee, and Shin]{LeeNIPS2018}
Kimin Lee, Kibok Lee, Honglak Lee, and Jinwoo Shin.
\newblock {A Simple Unified Framework for Detecting Out-of-Distribution Samples
  and Adversarial Attacks}.
\newblock \emph{Neural Information Processing Systems}, 2018.

\bibitem[Lee et~al.(2020)Lee, Ha, Zhang, and Kim]{Lee2020}
Soochan Lee, Junsoo Ha, Dongsu Zhang, and Gunhee Kim.
\newblock {A Neural Dirichlet Process Mixture Model for Task-Free Continual
  Learning}.
\newblock \emph{International Conference on Learning Representations}, 2020.

\bibitem[Li \& Hoiem(2017)Li and Hoiem]{Li2017}
Zhizhong Li and Derek Hoiem.
\newblock {Learning Without Forgetting}.
\newblock \emph{IEEE Transactions on Pattern Analysis and Machine
  Intelligence}, 40\penalty0 (12):\penalty0 2935--2947, 2017.

\bibitem[Liang et~al.(2018{\natexlab{a}})Liang, Heilmann, Gregory, Diallo,
  Carlson, Spell, Sigman, Roe, and Carin]{Liang2018a}
Kevin~J Liang, Geert Heilmann, Christopher Gregory, Souleymane~O Diallo, David
  Carlson, Gregory~P Spell, John~B Sigman, Kris Roe, and Lawrence Carin.
\newblock {Automatic Threat Recognition of Prohibited Items at Aviation
  Checkpoint with X-ray Imaging: A Deep Learning Approach}.
\newblock \emph{SPIE Anomaly Detection and Imaging with X-Rays (ADIX) III},
  2018{\natexlab{a}}.

\bibitem[Liang et~al.(2018{\natexlab{b}})Liang, Li, Wang, and
  Carin]{Liang2018b}
Kevin~J Liang, Chunyuan Li, Guoyin Wang, and Lawrence Carin.
\newblock {Generative Adversarial Network Training is a Continual Learning
  Problem}.
\newblock \emph{arXiv preprint arXiv:1811.11083}, 2018{\natexlab{b}}.

\bibitem[Lopez-Paz \& Ranzato(2017)Lopez-Paz and Ranzato]{Lopez2017}
David Lopez-Paz and Marc'Aurelio Ranzato.
\newblock {Gradient Episodic Memory for Continual Learning}.
\newblock \emph{Neural Information Processing Systems}, 2017.

\bibitem[Maddison et~al.(2017)Maddison, Mnih, and Teh]{MaddisonMT16}
Chris~J Maddison, Andriy Mnih, and Yee~Whye Teh.
\newblock {The Concrete Distribution: A Continuous Relaxation of Discrete
  Random Variables}.
\newblock \emph{International Conference on Learning Representations}, 2017.

\bibitem[McCloskey \& Cohen(1989)McCloskey and Cohen]{McCloskey1989}
Michael McCloskey and Neal~J Cohen.
\newblock {Catastrophic Interference in Connectionist Networks: The Sequential
  Learning Problem}.
\newblock \emph{The Psychology of Learning and Motivation}, 1989.

\bibitem[Michalowicz et~al.(2013)Michalowicz, Nichols, and
  Bucholtz]{kumarentropy}
Joseph~Victor Michalowicz, Jonathan~M. Nichols, and Frank Bucholtz.
\newblock \emph{Handbook of Differential Entropy}.
\newblock Chapman and Hall/CRC, 2013.
\newblock ISBN 1466583169.

\bibitem[Nalisnick \& Smyth(2016)Nalisnick and
  Smyth]{nalisnick2016stickbreaking}
Eric Nalisnick and Padhraic Smyth.
\newblock Stick-breaking variational autoencoders, 2016.

\bibitem[Nalisnick et~al.(2018)Nalisnick, Matsukawa, Teh, Gorur, and
  Lakshminarayanan]{nalisnick2018deep}
Eric Nalisnick, Akihiro Matsukawa, Yee~Whye Teh, Dilan Gorur, and Balaji
  Lakshminarayanan.
\newblock Do deep generative models know what they don't know?
\newblock \emph{arXiv preprint arXiv:1810.09136}, 2018.

\bibitem[Nguyen et~al.(2018)Nguyen, Li, Bui, and Turner]{Nguyen2018}
Cuong~V Nguyen, Yingzhen Li, Thang~D Bui, and Richard~E Turner.
\newblock {Variational Continual Learning}.
\newblock \emph{International Conference on Learning Representations}, 2018.

\bibitem[Parisi et~al.(2019)Parisi, Kemker, Part, Kanan, and
  Wermter]{Parisi2019}
German~Ignacio Parisi, Ronald Kemker, Jose~L. Part, Christopher Kanan, and
  Stefan Wermter.
\newblock {Continual Lifelong Learning with Neural Networks: {A} Review}.
\newblock \emph{Neural Networks}, 2019.

\bibitem[Ratcliff(1990)]{Ratcliff1990}
Roger Ratcliff.
\newblock {Connectionist Models of Recognition Memory: Constraints Imposed by
  Learning and Forgetting Functions}.
\newblock \emph{Psychology Review}, 1990.

\bibitem[Ribli et~al.(2018)Ribli, Horv{\'a}th, Unger, Pollner, and
  Csabai]{Ribli2018}
Dezs{\H{o}} Ribli, Anna Horv{\'a}th, Zsuzsa Unger, P{\'e}ter Pollner, and
  Istv{\'a}n Csabai.
\newblock {Detecting and Classifying Lesions in Mammograms with Deep Learning}.
\newblock \emph{Scientific reports}, 8\penalty0 (1):\penalty0 1--7, 2018.

\bibitem[Ritter et~al.(2018)Ritter, Botev, and Barber]{Ritter2018}
Hippolyt Ritter, Aleksandar Botev, and David Barber.
\newblock {Online Structured Laplace Approximations for Overcoming Catastrophic
  Forgetting}.
\newblock \emph{Neural Information Processing Systems}, 2018.

\bibitem[Rolnick et~al.(2019)Rolnick, Ahuja, Schwarz, Lillicrap, and
  Wayne]{Rolnick2019}
David Rolnick, Arun Ahuja, Jonathan Schwarz, Timothy Lillicrap, and Gregory
  Wayne.
\newblock {Experience Replay for Continual Learning}.
\newblock \emph{Neural Information Processing Systems}, 2019.

\bibitem[Rusu et~al.(2016)Rusu, Rabinowitz, Desjardins, Soyer, Kirkpatrick,
  Kavukcuoglu, Pascanu, and Hadsell]{Rusu2016}
Andrei~A Rusu, Neil~C Rabinowitz, Guillaume Desjardins, Hubert Soyer, James
  Kirkpatrick, Koray Kavukcuoglu, Razvan Pascanu, and Raia Hadsell.
\newblock {Progressive Neural Networks}.
\newblock \emph{arXiv preprint arXiv:1606.04671}, 2016.

\bibitem[Schwarz et~al.(2018)Schwarz, Luketina, Czarnecki, Grabska-Barwinska,
  Teh, Pascanu, and Hadsell]{Schwarz2018}
Jonathan Schwarz, Jelena Luketina, Wojciech~M Czarnecki, Agnieszka
  Grabska-Barwinska, Yee~Whye Teh, Razvan Pascanu, and Raia Hadsell.
\newblock {Progress \& Compress: A Scalable Framework for Continual Learning}.
\newblock \emph{International Conference on Machine Learning}, 2018.

\bibitem[Shin et~al.(2017)Shin, Lee, Kim, and Kim]{Shin2017}
Hanul Shin, Jung~Kwon Lee, Jaehong Kim, and Jiwon Kim.
\newblock {Continual Learning with Deep Generative Replay}.
\newblock \emph{Neural Information Processing Systems}, 2017.

\bibitem[Smith \& Gal(2018)Smith and Gal]{smith2018understanding}
Lewis Smith and Yarin Gal.
\newblock Understanding measures of uncertainty for adversarial example
  detection.
\newblock \emph{arXiv preprint arXiv:1803.08533}, 2018.

\bibitem[Teh et~al.(2007)Teh, Grür, and Ghahramani]{teh2007stick}
Yee~Whye Teh, Dilan Grür, and Zoubin Ghahramani.
\newblock Stick-breaking construction for the indian buffet process.
\newblock \emph{Artificial Intelligence and Statistics}, 2007.

\bibitem[van~de Ven \& Tolias(2018)van~de Ven and Tolias]{Van2018}
Gido~M van~de Ven and Andreas~S Tolias.
\newblock {Generative Replay with Feedback Connections as a General Strategy
  for Continual Learning}.
\newblock \emph{arXiv preprint arXiv:1809.10635}, 2018.

\bibitem[van~de Ven \& Tolias(2019)van~de Ven and Tolias]{vandeven2019three}
Gido~M van~de Ven and Andreas~S Tolias.
\newblock Three scenarios for continual learning.
\newblock \emph{arXiv preprint arXiv:1904.07734}, 2019.

\bibitem[Veniat et~al.(2021)Veniat, Denoyer, and Ranzato]{veniat2021efficient}
Tom Veniat, Ludovic Denoyer, and Marc'Aurelio Ranzato.
\newblock Efficient continual learning with modular networks and task-driven
  priors.
\newblock \emph{International Conference on Learning Representations}, 2021.

\bibitem[Williams(1992)]{williams1992simple}
Ronald~J Williams.
\newblock {Simple Statistical Gradient-Following Algorithms for Connectionist
  Reinforcement Learning}.
\newblock \emph{Machine learning}, 1992.

\bibitem[Xu \& Zhu(2018)Xu and Zhu]{xu2018reinforced}
Ju~Xu and Zhanxing Zhu.
\newblock {Reinforced Continual Learning}.
\newblock \emph{Neural Information Processing Systems}, 2018.

\bibitem[Yoon et~al.(2018)Yoon, Yang, Lee, and Hwang]{yoon2017lifelong}
Jaehong Yoon, Eunho Yang, Jeongtae Lee, and Sung~Ju Hwang.
\newblock {Lifelong Learning with Dynamically Dxpandable Networks}.
\newblock \emph{International Conference on Learning Representations}, 2018.

\bibitem[Zenke et~al.(2017)Zenke, Poole, and Ganguli]{Zenke2017}
Friedemann Zenke, Ben Poole, and Surya Ganguli.
\newblock {Continual Learning Through Synaptic Intelligence}.
\newblock \emph{International Conference on Machine Learning}, 2017.

\bibitem[Zhang et~al.(2019)Zhang, Sax, Zamir, Guibas, and Malik]{Zhang2019}
Jeffrey~O Zhang, Alexander Sax, Amir Zamir, Leonidas Guibas, and Jitendra
  Malik.
\newblock {Side-Tuning: Network Adaptation via Additive Side Networks}.
\newblock \emph{arXiv preprint arXiv:1912.13503}, 2019.

\end{thebibliography}
\end{@fileswfalse}
}

\begin{acronym}[CCCCCCCC]
    \acro{IBP}{Indian Buffet Process}
    \acro{MLP}{multilayer perceptron}
    \acro{SM}{supplemental materials}
    \acro{KL}{Kullback-Leibler}
\end{acronym}
\end{document}